\documentclass[10pt,twocolumn,letterpaper]{article}
\usepackage{iccv}
\usepackage{times}
\usepackage{epsfig}
\usepackage{graphicx}
\usepackage{amsmath}
\usepackage{amssymb}

\usepackage{amsfonts}
\usepackage{bbold}

\usepackage{multirow}
\usepackage{makecell}
\usepackage{subcaption}
\usepackage{verbatim}

\usepackage{multirow}
\usepackage{makecell}
\usepackage{subcaption}
\usepackage{verbatim}  

\usepackage[misc]{ifsym}

\usepackage[pagebackref=true,breaklinks=true,letterpaper=true,colorlinks,bookmarks=false]{hyperref}

\usepackage[breaklinks=true,bookmarks=false]{hyperref}

\iccvfinalcopy 

\def\iccvPaperID{****} 

\ificcvfinal\fi

\makeatletter
\newcommand{\printfnsymbol}[1]{%
  \textsuperscript{\@fnsymbol{#1}}%
}
\makeatother

\begin{document}

\title{Scale Attention for Learning Deep Face Representation: \\
A Study Against Visual Scale Variation}

\author{ Hailin Shi\textsuperscript{1}\thanks{Equal contribution. This work was performed at JD AI Research.}
~~~~ Hang Du\textsuperscript{1,2}\printfnsymbol{1} 
~~~~ Yibo Hu\textsuperscript{1}
~~~~ Jun Wang\textsuperscript{1} 
~~~~ Dan Zeng\textsuperscript{2}~~~~ Ting Yao\textsuperscript{1} 
\\
\textsuperscript{1}JD AI Research ~~~~ \textsuperscript{2}Shanghai University 
\\
{\tt\small hailinshi.work@gmail.com~~~~ duhang@shu.edu.cn}
}

\maketitle
\ificcvfinal\thispagestyle{empty}\fi

\begin{abstract}
Human face images usually appear with wide range of visual scales.
The existing face representations pursue the bandwidth of handling scale variation via multi-scale scheme that assembles a finite series of predefined scales.
Such multi-shot scheme brings inference burden, and the predefined scales inevitably have gap from real data.
Instead, learning scale parameters from data, and using them for one-shot feature inference, is a decent solution.
To this end, we reform the conv layer by resorting to the scale-space theory, and achieve two-fold facilities: 1) the conv layer learns a set of scales from real data distribution, each of which is fulfilled by a conv kernel; 2) the layer automatically highlights the feature at the proper channel and location corresponding to the input pattern scale and its presence.
Then, we accomplish the hierarchical scale attention by stacking the reformed layers, building a novel style named SCale AttentioN Conv Neural Network (\textbf{SCAN-CNN}). 
We apply SCAN-CNN to the face recognition task and push the frontier of SOTA performance.
The accuracy gain is more evident when the face images are blurry.
Meanwhile, as a single-shot scheme, the inference is more efficient than multi-shot fusion.
A set of tools are made to ensure the fast training of SCAN-CNN and zero increase of inference cost compared with the plain CNN.
\end{abstract}

\begin{figure}[!t]
\centering
\includegraphics[scale=0.3]{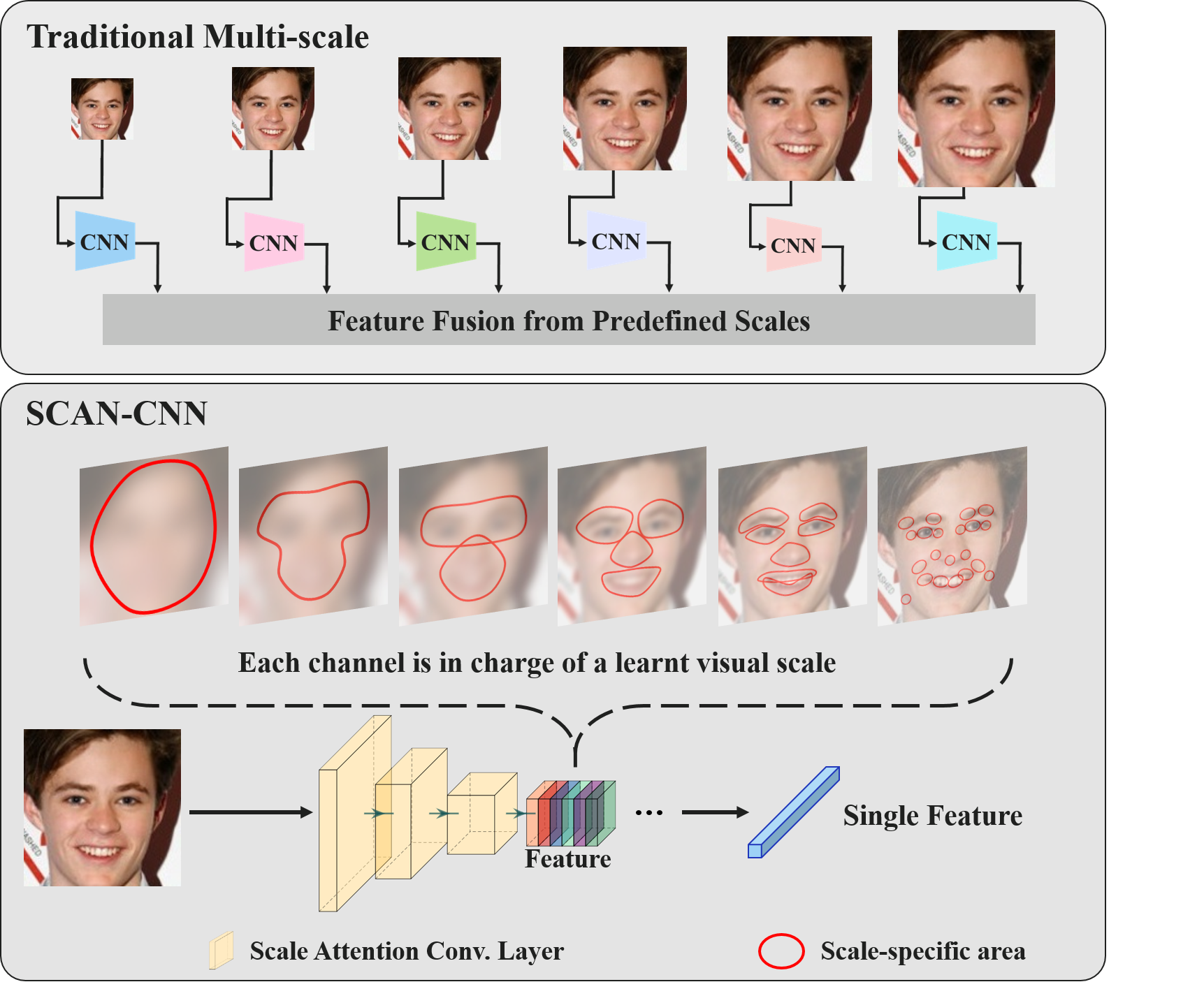}
\caption{ 
The traditional multi-scale practice employs multiple CNNs to compute the features from a series of input images of predefined scales, and fuses the features for recognition. 
In contrast, SCAN-CNN 
\textit{\textbf{learns the scale parameter from real data, and highlights the feature at the channel corresponding to the scale of the pattern, and at the spatial location where the pattern presents}}.
}
\label{multi_scale}
\vspace{-10pt}
\end{figure}

\section{Introduction}
\label{sec:intro}

Visual scale is an inherent property of real-world objects (including human face).
The concept of visual scale is not only related to object size and image resolution, but also closely related to imaging quality~\cite{lindeberg1996scale}.
Given a face in image, its visual scale could vary remarkably from fine to coarse along with the increase of blurring degree.
Spanning a wide range of \textbf{scale variation across face images}, this issue widely exists in face datasets.
For instance, the faces captured from video are more blurry than the still photos, which is an adverse factor to face representation learning. 
The multi-scale representation is an usual way to enlarge the bandwidth of handling scale variations. The existing methods~\cite{taigman2014deepface,sun2014deep,sun2014deepid2,liu2017sphereface,wang2018cosface} achieve this goal by fusing the features from a series of predefined scales, and gains certain robustness to scale variation. 
Such scheme, however, leads to the suboptimal accuracy of face recognition, since a finite range of scales are predefined empirically, without considering the scale distribution of real data; besides, the multi-shot computation throughout the multiple scales, each of which requires a complete forward propagation, leads to the inefficient inference. 
One may expect that designing \textbf{attention mechanism related to real-data visual scale} is a decent solution to replace the multi-scale scheme.




Recently, the attention concept has been widely studied in computer vision. 
Many approaches~\cite{Wang2017ResidualAN,Hu2018SqueezeandExcitationN,woo2018cbam,chen2017sca,yu2018learning,fu2019dual,wang2020eca} develop the spatial and channel attention schemes to exploit the effective information from data, and thereby promote the representation learning for visual tasks. 
Among them, Wang~\etal~\cite{Wang2017ResidualAN} propose a residual attention network for leveraging the spatial attention.
SENet~\cite{Hu2018SqueezeandExcitationN} designs a squeeze-and-excitation module to fulfil the channel-wise attention.
Besides, CBAM~\cite{woo2018cbam} infers both spatial and channel attention to refine the feature maps. 
As to deal with the scale variation, some studies~\cite{chen2016attention,chu2017multi} adopt the feature fusion with multiple scales, and others~\cite{zhang2017scale,li2019selective} adjust the receptive field sizes of convolution kernels and aggregate the different levels of feature contexts. 
These methods, however, overlook the attention mechanism related to visual scale, whereas the scale attention shows its necessity in the above discussion.
Therefore, we aim at modeling the scale attention and facilitating the CNNs to highlight the feature at the optimal scale and location subject to the input facial pattern.
The \textbf{basic idea} is that: (1) in training phase, learning a small set of parameters from real data, which can depict the real visual scale; (2) in inference phase, utilizing the learnt scale parameters for establishing an attention mechanism related to the input scale.


In this study, we accomplish the modeling of visual scale attention (\textbf{SCAN}) by resorting to the scale-space theory~\cite{lindeberg1990scale,lindeberg1998feature}, and implement the scale attention with convolutional neural network, named \textbf{SCAN-CNN}. 
Specifically, we first build the scale-space representation by equipping the conv layer with a learnable Gaussian kernel, regarding the plain convolution kernel as the role of pattern signature in the scale-space theory.
Then, given the non-monotonic property of the normalized derivative in scale-space representation~\cite{lindeberg1998feature}, the automatic scale selection is fulfilled as an objective for learning scale parameters.
Therefore, in the data-driven manner, SCAN-CNN learns the scale parameter subject to the real data.
Finally, in inference, SCAN-CNN employs the learnt scale parameter to highlight the features at the right channel and right location corresponding to the input and its scale, achieving the attention related to visual scale.
We equip the scale attention to every conv layer to exploit the large receptive field in deep layers, so the kernels in deep layers can perceive a large range of scales.

Compared with the traditional multi-scale methods, \textbf{the advantages of SCAN-CNN are in two folds}. 
\textbf{(1)} The scale parameters are learnt from real data, which depict more precise visual scale than the predefined ones for characterizing the real data.
\textbf{(2)} Resorting to the scale attention, SCAN-CNN infers the feature that emphasized at the channel corresponding to the pattern scale, and at the location where the pattern presents;
by doing so, SCAN-CNN can yield scale-specific features in a single-shot manner, rather than multi-shot computation in mult-scale scheme, 
simultaneously promoting the accuracy and efficiency.

Compared with plain CNNs, \textbf{the advantage of SCAN-CNN} is that, it can infer scale-specific representation for better robustness to scale variation, owing to the explicit modeling of visual scale attention into CNN.
Meanwhile, for inference, SCAN-CNN does not increase model parameter and computation cost, as the Gaussian kernel and scale parameter can be absorbed into existing conv kernel once learnt.

It is noteworthy that, SCAN-CNN has its potentials for general visual tasks; whereas, in this paper, we apply it and verify its effectiveness to the face recognition task.
We conduct comprehensive experiments to show the improvement brought by SCAN-CNN on various benchmarks, including on the blurry face recognition.
Based on these results, we wish this work will shed light on more visual tasks. 


The contribution of this paper is summarized as follows:
 \begin{itemize}
     \item We propose SCAN-CNN to accomplish the attention mechanism related to the real data visual scale, boosting the face representation with better robustness to scale variation.
     \item Apart from the accuracy gain, SCAN-CNN also achieves better inference efficiency compared with traditional multi-scale routine, and zero increase of inference cost compared with the plain CNNs.
     \item Based on SCAN's flexibility with many architectures and loss functions, we conduct comprehensive experiments, and push the frontier of state-of-the-art (SOTA) performance on various face recognition benchmarks, including on the blurry cases.  
 \end{itemize}

\section{Related Work}
\label{sec:related}

\subsection{Attention Mechanism} 
\label{sec:related:attention}
The attention mechanism is derived from the human visual system which considers more about the informative components of the whole scene~\cite{itti2001computational}. 
Recently, the attention mechanism has achieved great success in many visual tasks~\cite{Wang2017ResidualAN,Hu2018SqueezeandExcitationN,woo2018cbam,chen2017sca,yu2018learning,fu2019dual,wang2020eca}. 
Among them, Wang~\etal~\cite{Wang2017ResidualAN} propose a residual attention network for leveraging the spatial attention to refine the feature maps and guide the representation learning. 
Hu~\etal~\cite{Hu2018SqueezeandExcitationN} present a squeeze-and-excitation module to effectively compute the channel attention for CNNs. 
Besides, ECA-Net~\cite{wang2020eca} designs an efficient channel attention module to learn the effective channel attention with low model complexity.
In addition, some of them~\cite{chen2017sca,woo2018cbam,fu2019dual} explore to combine the spatial attention and channel attention together to further improve the representation ability of CNNs. The above methods mainly focus on designing attention modules to infer the spatial or channel attention for better feature representation learning.

\subsection{Multi-scale Representation}
\label{sec:related:multiscale}
Scale variation is one of the key challenges in many vision tasks.
To address this challenge, an usual way is to exploit the multi-scale representation from the image pyramid or feature pyramid. 
For example, some methods~\cite{sun2014deep,sun2014deepid2} compute the multi-scale face features by feeding the multiple facial patches to the network ensemble.
FPN~\cite{lin2016feature} proposes a top-down architecture with skip connections to aggregate the high- and low-level features. 
Some approaches~\cite{chen2016attention,chu2017multi,zhang2017scale,li2019selective,wang2019salient} aim to adaptively fuse the features of multiple scales. 
Among them, Chen~\etal~\cite{chen2016attention} pass the multi-scale input images to the network and exploit attention models to aggregate multi-scale features. 
SK-Net~\cite{li2019selective} proposes the selective kernel convolution to adjust its receptive field sizes and aggregate the different levels of feature contexts. 
These methods, however, take into account a finite range of predefined scales, which leads to the suboptimal accuracy and inefficient inference. 
Differently, we explicitly model the visual scale into CNNs, and fulfil the scale attention to boost the discriminative features. To the best of our knowledge, our method is the first attempt towards the scale attention for face recognition. 

\section{Preliminary of Scale-space Theory}
\label{sec:scalespace}
The scale-space theory is a long-standing scheme for visual representation computing. In this section, we revisit the Gaussian scale-space representation~\cite{lindeberg1990scale} and the automatic scale selection~\cite{lindeberg1998feature}, facilitating the following section to present our method that generalizes the scale-space theory to the learning-based regime.

\subsection{Scale-space Representation}
\label{sec:scalespace:repre}

The notion of scale is essential to computing visual representation.
Lindeberg~\cite{lindeberg1990scale} proposes that the observed signal can be regarded as the initial state of ``temperature'' in space, then the scale-space representation can be described by the heat equation of isotropic diffusion, in which the ``time'' variable corresponds to the visual scale\footnote{One can refer to the details in the supplementary material}. We can obtain the linear solution in closed form by the Gaussian kernel convolution.
Given the observed signal $f(x)$ with $D$-dimensional coordinates $x = (x_1, \dots, x_D)^T$, the scale-space representation $F(x; t)$ is formulated as 
\begin{equation}
F(\cdot; t) = g(\cdot; t) \circledast f(\cdot), 
\label{scale_space_representation}
\end{equation} 
where $t \in \mathbb{R}_+$ indicates the scale, $\circledast$ denotes the convolution operation, and $g$ is the Gaussian kernel with the variance associated with the scale $t$,
\begin{equation} 
g(x; t)= {\frac{1}{(2 \pi t)^{D/2}} e^{-\left(x^{2}_1 + \dots +x^{2}_D \right) /(2t)}} . 
\label{gaussian_kernel}
\end{equation}

The initial condition is given by $F(\cdot; 0) = f(\cdot)$, which preserves the maximum spatial frequency (\textit{aka.} the finest scale). With the increase of scale $t$, the scale-space representation gradually loses the high frequency information, meanwhile the structures of coarse scale become dominant.

\begin{figure}[tp]
\centering
\includegraphics[scale=0.3]{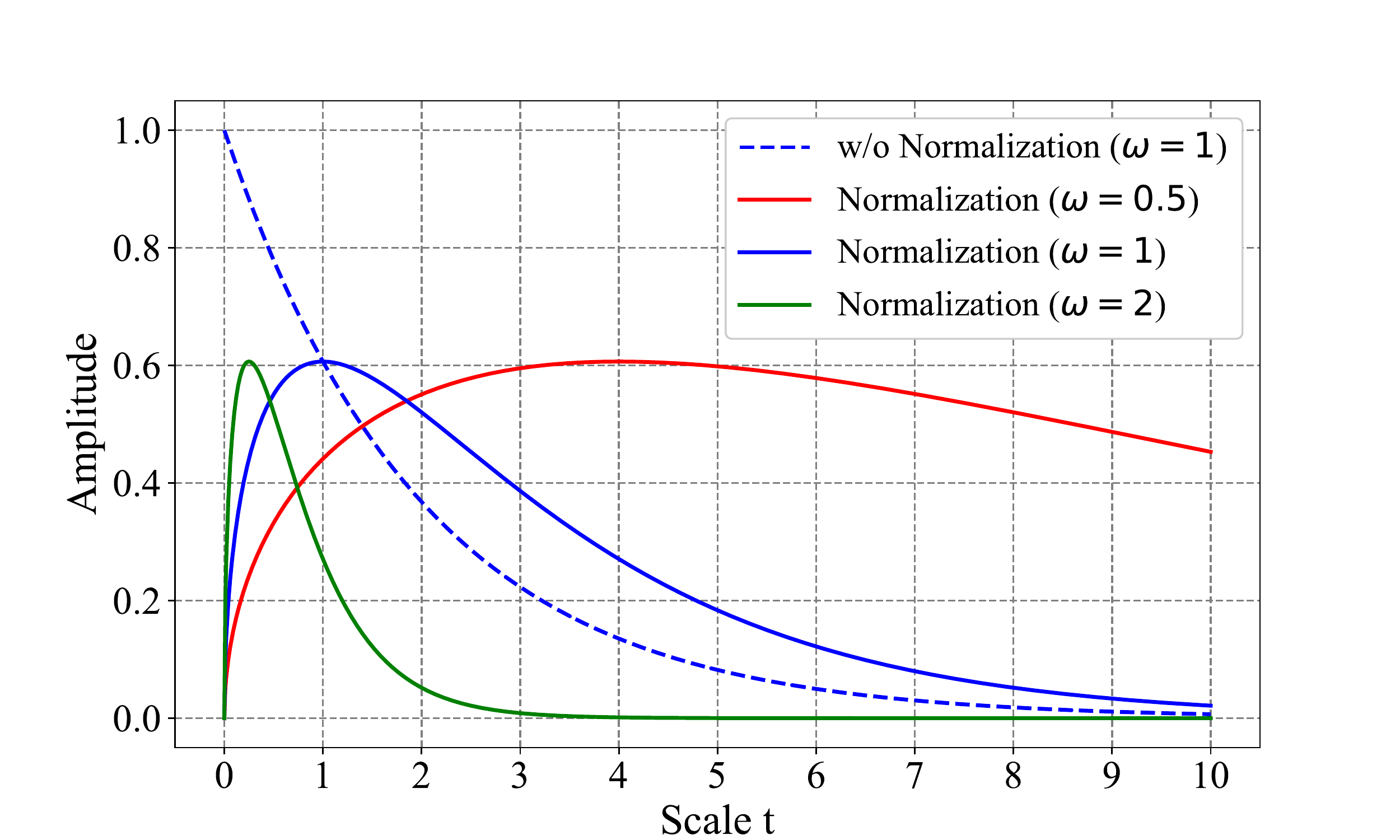}
\caption{The amplitude of the ordinary and the normalized ($\gamma=1$) derivatives as a function of scale. The dashed curve represents the ordinary derivative without normalization, whose amplitude monotonically decreases. The solid curves are the amplitudes of the normalized derivatives, which attain their maximum at their respective scale (in inverse proportion to the signal frequency).
}
\label{fig_sin}
\vspace{-1em}
\end{figure}

\subsection{Automatic Scale Selection} 
\label{sec:scalespace:autsel}
As presented above, the scale-space representation gradually focuses on the coarse scale with the increase of $t$. In other words, as the high-frequency portion monotonically diminished, the increasing $t$ should not create new local extremum in the representation.
Without loss of generality, we take a simple example of $f(x) = sin(\omega x)$. Its scale-space representation is computed as $F(x;t) = e^{-\omega^2 t/2} sin(\omega x)$. 
We can derive the amplitude of the derivatives of any $m$th order in the scale-space representation as $F_{x^m, max}(t) = \omega^m e^{-\omega^2 t/2}$, which decreases exponentially with the scale (the dashed curve in Fig.~\ref{fig_sin}). This monotonic property indicates that the original scale-space representation is not capable for the automatic scale selection. It is critical to formulate a non-monotonic function, resulting in a peak at which $t$ is proportional to the signal scale.


As an explicit scheme for scale selection, the normalized derivative~\cite{lindeberg1998feature} in scale-space representation is proposed by reshaping the ordinary derivative operator to $\partial_{\xi} = t^\gamma \partial_x $, where $\gamma$ is a hyperparameter for modulating the normalization. 
Then, the ($m$th order) normalized derivative of the sinusoid signal in scale-space representation behaves with the amplitude of $F_{\xi^m, max}(t) = t^{m\gamma} \omega^m e^{-\omega^2 t/2}$ (the solid curves in Fig.~\ref{fig_sin}). We can observe that, compared with the ordinary derivative, the normalized derivative has two valuable properties. 
First, the response is a non-monotonic function over scale, and its maximum is unique globally when setting $\gamma=1$.
Second, the unique maximum is attained at the scale of $\hat t = \gamma m / {\omega^2}$, which precisely reflects the real scale of the signal (\ie proportional to the signal wavelength $2 \pi / \omega$).
Based on the two-fold advantages, the normalized derivative can be suitably applied for the automatic scale selection in the scale-space representation.

\section{The Proposed Approach}
\label{sec:method}
\subsection{SCAN-CNN Modeling} 
\label{sec:method:modeling}
Benefiting from the properties of the normalized derivative in scale-space representation, we can easily identify the real scale of the signal by finding the maximum over scales and its corresponding $\hat t$.
Such ability of scale selection can support the modeling of scale attention and promote the deep face representation against scale variation.
Yet, the remaining challenge consists in how to implement the scale-space representation and the automatic scale selection in a learning style.
We dwell on two facts that:
(1) the sinusoid signal is taken as an example in the above analysis, while the real-world signals can be decomposed by the linear combination of the sinusoidal family, which means the scale-space theory is valid for the processing of practical signals, such as 2D images;
(2) the local feature descriptor, or so-called pattern signature in literature, or equally the convolution kernel of CNN, can be decomposed by the linear combination of the finite discrete operators of derivatives of different orders.

\noindent
\textbf{Scale attention mechanism.}
Therefore, the scale attention in CNN can be accomplished in two steps.
The basic idea is to reform each convolution operation in CNN.
First, given the input 2D image $f_{in}(x)$ with $x = (x_1, x_2)^T$ and the kernel $k(x)$ of a conv layer, the convolution is performed on the scale-space representation of $f_{in}(x)$.
The scale-space representation is computed by $g(\cdot; t) \circledast f_{in}(\cdot)$ according to Eq.~\ref{scale_space_representation}, and the Gaussian kernel $g(\cdot; t)$ is an instance of Eq.~\ref{gaussian_kernel} in 2D form.
Thus, the output of this layer changes from $f_{out}(\cdot)=k(\cdot) \circledast f_{in}(\cdot)$ to
\begin{equation}
f_{out}(\cdot) = k(\cdot) \circledast g(\cdot; t) \circledast f_{in}(\cdot).
\label{primary_output}
\end{equation}
Second, we regard the kernel $k$
as the linear combination of the finite discrete operators of the ordinary derivatives, and reshape it to the normalized derivatives $t^{\gamma_m} k$, where $\gamma_m = \gamma \cdot m$ and $m$ is the order of $k$. Finally, the output can be written as
\begin{equation}
f_{out}(\cdot) = t^{\gamma_m} k(\cdot) \circledast g(\cdot; t) \circledast f_{in}(\cdot).
\label{final_output}
\end{equation}

By Eq.~\ref{final_output}, the conv layer is empowered to infer the scale-space representation, as well as the normalized derivatives. 
Note that the scale parameter $t$ is learnable, and the derivative operator $k$ is exactly the original convolution kernel which is also learnable. 
For the simplicity, we set $\gamma_m$ as a constant one in the experiments and find it performs stably well.
In summary, $k$ is in charge of learning the visual descriptor corresponding to the input pattern, meanwhile $t$ is in charge of learning the visual scale of the input pattern.
A conv layer owns a set of kernels, that is, a set of $t$'s and $k$'s.
Once $t$'s and $k$'s are learnt jointly along with the network training, a range of real-data visual scales are learnt from training set.
Then, this conv layer is able to emphasize the feature response of the input pattern at the corresponding channel and the corresponding location through the attained maximum of the normalized derivatives.
In this way, the scale attention mechanism is established for a single conv layer.

\noindent
\textbf{Hierarchical SCAN-CNN.}
With the above practice, we equip all the conv layers in CNN with scale attention, and thereby complete the modeling of hierarchical scale attention. 
The purpose of such hierarchical implementation is to take advantage of the \textbf{large receptive field in deep layers, which enables the small kernels (mostly $3\times3$) to depict the coarse scales} that covers large spatial size.
The baseline CNN can be any choice of the existing architectures, of which the conv layers are reformed by Eq.~\ref{final_output}.
Note that the output of each layer is activated by $ReLU(f_{out})$, and the activation function is omitted in Eq.~\ref{final_output} for simplicity.

\noindent
\textbf{Model weight.}
In the training stage, the additional parameter only comes from $t$, since $k$ is the original kernel of plain CNN. Considering each $k$ is associated with an $t$, SCAN-CNN merely increases a small number of parameters in training. For example, if the kernel size of $k$ is $3\times3$, then the additional parameter number is one-ninth of the baseline CNN.

\noindent
\textbf{Efficient inference.}
In the inference stage, the operators in scale attention conv layer can be absorbed as a single kernel $k_s(\cdot) = t^{\gamma_m} k(\cdot) \circledast g(\cdot; t)$ in advance. Then, the inference output is computed as \begin{equation}
f_{out}(\cdot) = k_s(\cdot) \circledast f_{in}(\cdot),
\label{final_efficient_output}
\end{equation}
which does not require any extra cost compared with the regular conv layer. 
Correspondingly, SCAN-CNN has the same inference efficiency with the baseline CNN.

\subsection{Training Supervision}   
\label{sec:method:training}
The training objective is critical to SCAN-CNN, since the scale attention needs the correctly-learnt scale parameter as element (see Sec.~\ref{sec:exp:ablation}). 
The property of normalized derivative shows that the real visual scale is estimated by $\hat t$, at which the maximum response is attained. Therefore, it is necessary to design a learning objective that maximizes the output of the scale attention conv layer for optimizing the scale parameter. The response-maximization objective is formulated explicitly by 
\begin{equation}
\mathcal{L}_{scale} = exp({-\frac{\lambda ||f _{out}||_{2}  }{H \cdot W}} ), 
\label{loss_scale}
\end{equation}
where $H$ and $W$ are the height and width of $f_{out}$, respectively, and $\lambda$ is a hyperparameter. The total training loss is the combination of the response-maximization objective and the common loss function of face recognition,
\begin{equation}
\mathcal{L}_{total} = \mathcal{L}_{scale} + \mathcal{L}_{rec}. 
\label{loss_total}
\end{equation}
$\mathcal{L}_{rec}$ can be any common loss function of face recognition, such as Arc-softmax~\cite{deng2019arcface}, AdaM-softmax~\cite{liu2019adaptiveface}, \etc, and SCAN-CNN stably performs well with them (Fig.~\ref{loss_function}).

\noindent
\textbf{Efficient training.}
It is worth noting that the response-maximization objective only takes effect for the corresponding layer.
Therefore, the naive implementation of  backward-propagation in PyTorch is inefficient in this case. We address this issue by detaching the corresponding nodes from the computing graph and thus cutting off the redundant gradient flow.

\noindent
\textbf{Local shortcut connection.}
In practice, the response-maximization objective tends to cause the degeneration in most operators whose $t$ collapses to near zero (Fig.~\ref{distribution_c}). 
To deal with this issue, we implement the SCAN-CNN with the local shortcut connection.
Specifically, a regular convolution with an independent kernel $k_i$ is concurrently performed within the scale attention conv layer,
\begin{equation}
f_{out} = k_s \circledast f_{in} + k_i \circledast f_{in}
= (k_s + k_i) \circledast f_{in}
= k_u \circledast f_{in}.
\label{final_efficient_shortcut_output}
\end{equation}
Such residual structure helps SCAN-CNN to prevent the degradation (Table~\ref{ablation}).
Moreover, we can absorb $k_i$ and $k_s$ as an ultimate kernel $k_u$ to keep the efficient inference.

\section{Discussion on Scale Attention}
\label{sec:disc}
\noindent
\textbf{Visual scale concept.} Visual scale is closely related to, but not equal to image resolution and object size. For instance, given a face in image (fixed size and fixed resolution), its visual scale varies from fine to coarse along with the image degraded from clear to blurry, and accordingly $t$ increases.

\noindent
\textbf{Compare with traditional multi-scale.}
First, rather than the naive ensemble of a series of predefined visual scales, the core concept of SCAN-CNN is to learn the visual scales from data distribution, and use them for feature inference.
Compared with presetting the input scale, the learnt scale parameters are more suitable to depict the real data visual scale.
Second, SCAN-CNN improves the inference efficiency with one-shot scheme, compared with the multi-shot computation in multi-scale.


\noindent
\textbf{Compare with traditional CNNs.}
First, the plain CNNs, although having certain multi-scale capacity, infer the feature in a scale-agnostic style, which is suboptimal. 
Instead, SCAN-CNN establishes scale attention mechanism that \textit{\textbf{the maximum response is attained at the channel whose $t$ corresponds to the visual scale of input pattern, and at the location of presence of the pattern}}. 
Resorting to the scale attention mechanism, 
SCAN-CNN can raise the usage and efficacy of each convolution kernel, yielding scale-specific features and thereby better robustness to scale variation.
Second, SCAN-CNN does not increase inference cost compared with the counterpart CNN, since the scale attention operator can be absorbed into the conv kernel once learnt.

\noindent
\textbf{Attention mechanism with learnt $t$.}
In SCAN-CNN, although $t$, which is a parameter rather than a response, is fixed once learnt, the response is attained maximum at the channel corresponding to the input pattern scale, and at the location of its presence.
We should notice that $t$ plays the role such like the learnt parameters in AttentionNet~\cite{woo2018cbam} and SENet~\cite{Hu2018SqueezeandExcitationN} that in charge of enlarging the response in the attended region.
For example, given two channels $c_1$ and $c_2$ with their respective scale parameters $t_1$ and $t_2$ that $t_1 > t_2$, then a coarse pattern's feature response is highlighted at the location $(w_1, h_1, c_1)$, and a fine pattern's feature is highlighted at $(w_2, h_2, c_2)$. 
Note that $(w_1, h_1)$ and $(w_2, h_2)$ are the spatial locations of the presence of coarse pattern and fine pattern, respectively.

\noindent
\textbf{The necessity of scale attention to face.}
There are two major reasons for applying scale attention to face recognition.
First, in practice, face images are often captured by video cameras, leading to serious blurring issue, and thus large visual scale variation. In this case, SCAN-CNN shows great advantage to deal with the blurred test sets and real-world surveillance scenario (Sec.~\ref{sec:exp:compare_plain}).
Second, the facial component scale variation is another overlooked issue. For instance, eyes and mouth always appear at finer scale than the entire face. 
Given the same computational budget, SCAN-CNN yields features that are more scale-specific than the plain counterpart. This is why SCAN-CNN still performs better when the test sets are not blurred (Sec.~\ref{sec:exp:compare_plain}).


\noindent
\textbf{Dependence on receptive field.}
Although the convolution kernel size is usually small ($3\times3$), the receptive field becomes large as the increase of network depth. This property enables the small kernels in deep layer to capture coarse scale from a wide scope of input.

\section{Experiment}
\label{sec:exp}

\subsection{Datasets and Experimental Settings}
\label{sec:exp:setting}
\noindent\textbf{Training data.} We conduct the experiments with three training datasets, including the cleaned CASIA-WebFace~\cite{yi2014learning}, MS1M-v1c~\cite{trillionpairs.org} 
, and QMUL-SurvFace~\cite{cheng2018surveillance}. 
Specifically, CASIA-WebFace is employed in the ablation study, while MS1M-v1c and QMUL-SurvFace are utilized in the comparison with SOTA methods. 
For a strict and precise evaluation, we follow the lists of~\cite{wang2019co,Wang2019MisclassifiedVG} to remove the overlapping identities between the training datasets and test datasets. 

\noindent\textbf{Test data.} To thoroughly evaluate the effectiveness of SCAN-CNN, we test the models on ten benchmarks, including LFW~\cite{huang2008labeled}, BLUFR~\cite{liao2014benchmark}, AgeDB-30~\cite{moschoglou2017agedb}, CFP-FP~\cite{sengupta2016frontal}, CALFW~\cite{zheng2017cross}, CPLFW~\cite{zheng2018cross}, MegaFace~\cite{kemelmacher2016megaface}, IJB-B~\cite{Whitelam2017IARPAJB}, IJB-C~\cite{Maze2018IARPAJB}, and QMUL-SurvFace~\cite{cheng2018surveillance}. These test datasets contain various facial variations, including large pose, large age gap, motion blur, low resolution~\textit{etc}.

\noindent\textbf{Data prepossessing.} 
The facial region is detected from images by FaceBoxes~\cite{zhang2017faceboxes}. Then, all the faces are aligned by five facial landmarks and cropped to 144$\times$144 RGB images. The pixel values are normalized into $[-1, 1]$.

\begin{table}[t]
\begin{center}
\centering
\caption{The ablation study (\%) on the three components of SCAN-CNN, \ie SAC, LSC and RMO. The full assembly achieves the best results, indicating all the components are necessary to SCAN-CNN.
}
\label{ablation}
\resizebox{\linewidth}{!}{
\begin{tabular}{|c|c|c||c|c|c|c|c|c|}
\hline
\multirow{2}{*}{SAC}&
\multirow{2}{*}{RMO}&
\multirow{2}{*}{LSC}&
\multirow{2}{*}{AgeDB}&
\multirow{2}{*}{CFP}&
\multirow{2}{*}{CALFW}&
\multirow{2}{*}{CPLFW}&
\multicolumn{2}{c|}{BLUFR}
\\\cline{8-9} &&&&& && 
$10^{-4}$ & 
$10^{-5}$ \\
\hline\hline 
-&-&-&87.62&90.14&86.08&76.10&83.67&65.19\\
\hline 
\checkmark &-&-&88.00&90.80&85.55&76.23&82.32&64.79\\
\hline 
\checkmark &\checkmark &- &88.23&90.66&85.92&76.98&82.11&64.23\\
\hline 
\checkmark &- &\checkmark 
&87.57&90.34&84.98&75.17&79.95&62.67\\
\hline 
\checkmark &\checkmark &\checkmark 
&\textbf{89.63}&\textbf{91.27}&\textbf{86.78}&\textbf{77.73}&\textbf{85.75}&\textbf{68.48}\\
\hline
\end{tabular}}
\end{center}
\vspace{-20pt}
\end{table}

\noindent\textbf{CNN architecture and loss function.} SCAN enables the flexible combination with various network architectures and loss functions. 
In the ablation study, ResNet-18~\cite{he2016deep} is adopted as the backbone to validate the effectiveness of scale attention. 
In the comparison experiments, more complicated networks are involved, including ResNet-50 and -101~\cite{he2016deep}, SE-ResNet-50 and -101~\cite{Hu2018SqueezeandExcitationN}, and Attention-56~\cite{wang2017residual}.
Moreover, SCAN-CNN is validated with seven loss functions, including softmax, A-softmax~\cite{liu2017sphereface}, AM-softmax~\cite{wang2018cosface}, MV-AM-softmax~\cite{Wang2019MisclassifiedVG},  Arc-softmax~\cite{deng2019arcface}, Adacos~\cite{zhang2019adacos}, and AdaM-softmax~\cite{liu2019adaptiveface}.

\noindent\textbf{Training and evaluation.} 
We set the batch size as 256 on CASIA-WebFace dataset. 
The learning rate starts with 0.1 and is divided by 10 at 16, 24, 28 epoch, and the training is finished at 30 epochs. 
On MS1M-v1c dataset, the batch size is 128. The learning rate starts with 0.1 and is divided by 10 at 5, 9, 11 epoch, and the training is terminated at 12 epochs.
On QMUL-SurvFace dataset, the batch size is 128. The learning rate starts with 0.01 and is divided by 10 at 12, 16 epoch, and the training is finished at 20 epochs. 
We set the momentum to 0.9 and the weight decay to $10^{-4}$. 
In the evaluation stage, we extract 512-dimension features from the last fully connected layer of the backbone as the face representation, and utilize the cosine similarity as the similarity metric.

\noindent\textbf{More details in SCAN-CNN.}
The scale attention is applied to per filter in every conv layer of SCAN-CNN.
Thus, each filter has its own $t$.
For example, when the backbone is chosen as ResNet-50, then each of its kernel in every conv layer is equipped with scale attention, and the SCAN-CNN is called SCAN-ResNet-50.
In consideration of edge effect, given the kernel size of $k$ is $3\times3$, we set $g$ as $5\times5$ to keep the size of $k \circledast g$ unchanged as $3\times3$.
The scale parameter $t$ is initialized as 1, and updated without the constraint of weight decay.
Both $\gamma_m$ and $\lambda$ are fixed as 1.

\subsection{Ablation Study}
\label{sec:exp:ablation}
\begin{figure}[t]
\centering
\begin{minipage}[t]{0.4\linewidth}
\includegraphics[height=2.5cm]{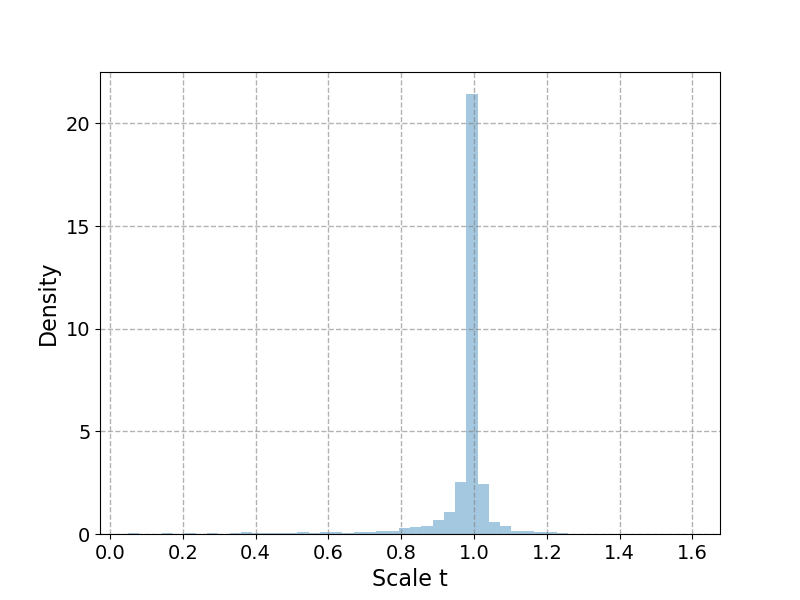}
\subcaption{SAC}
\label{distribution_a}
\end{minipage}
\hspace{10pt}
\begin{minipage}[t]{0.4\linewidth}
\includegraphics[height=2.5cm]{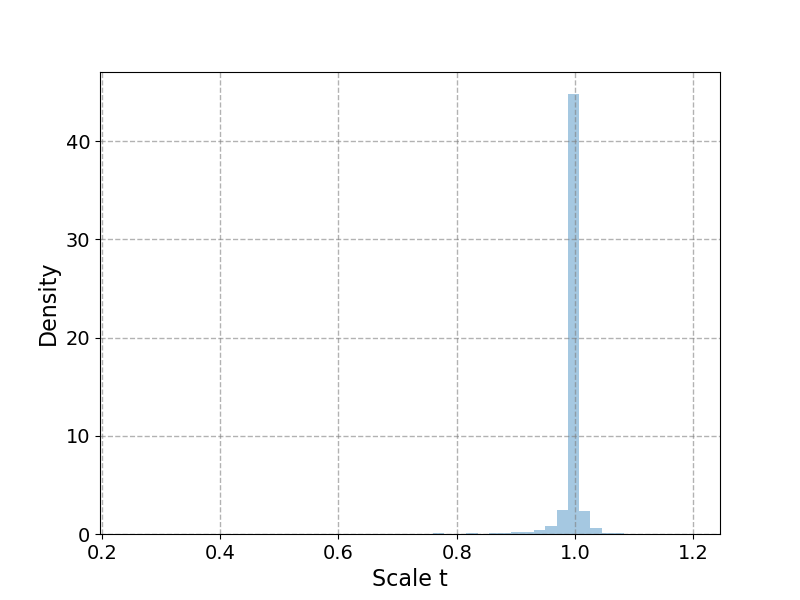}
\subcaption{SAC+LSC}
\label{distribution_b}
\end{minipage} 
\\
\begin{minipage}[t]{0.4\linewidth}
\includegraphics[height=2.5cm]{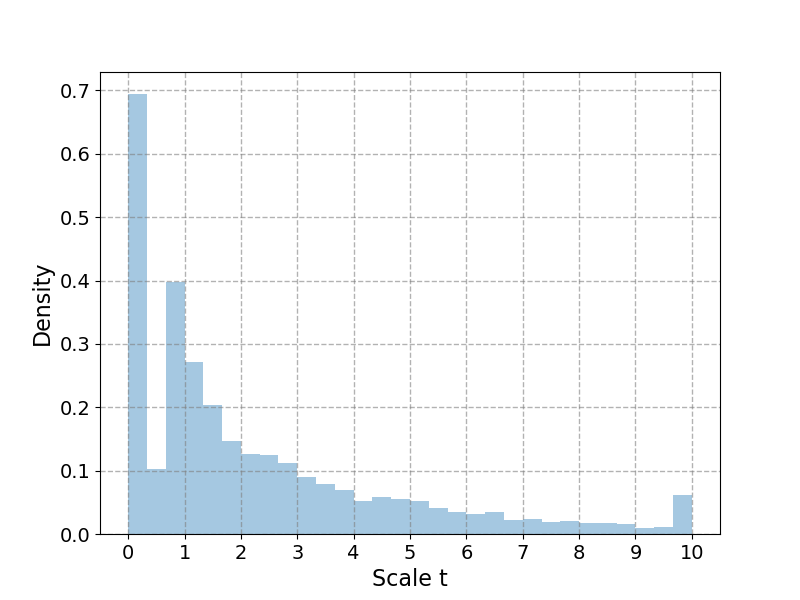}
\subcaption{SAC+RMO}
\label{distribution_c}
\end{minipage}
\hspace{10pt}
\begin{minipage}[t]{0.4\linewidth}
\includegraphics[height=2.5cm]{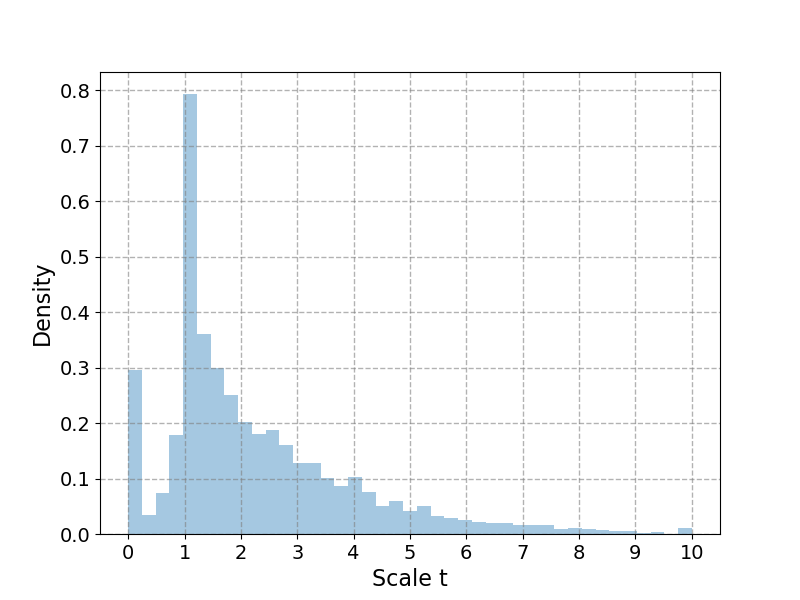}
\subcaption{SAC+LSC+RMO}
\label{distribution_d}
\end{minipage}
\caption{The distribution density of $t$ reflects the necessity of the model components.
(a) The solo version of SAC layer.
(b) The SAC layer with LSC. 
(c) The SAC layer trained with RMO.
(d) The full assembly of SAC, LSC and RMO, empowering $t$ to correctly learn the visual scale from real data distribution.
}
\label{scale_density}
\vspace{-1em}
\end{figure}

\noindent\textbf{Ablation of model components.} 
At the beginning of experiment, it is important to verify the effect of SCAN-CNN's three major components, which are scale attention conv (SAC) layer, local shortcut connection (LSC) and response-maximization objective (RMO).
So, we check the performance gain by each of them. 
As we can see in Table~\ref{ablation}, firstly, the solo version of SAC layer can obtain improvements compared with the baseline in most terms.
Secondly, SAC layer trained with RMO results in steady increases. 
Thirdly, in the bottom row, the full assembly of SAC, LSC and RMO achieves the best result by all metrics. 
To explore the reasons, we plot the density distribution of scale parameter $t$ under these settings.
In Fig.~\ref{scale_density} (a) and Fig.~\ref{scale_density} (b), we can observe that $t$ gathers around the initial value, implying the scale learning fails. 
This is because the lack of explicit learning objective for pursuing the real visual scale.
By the joining of RMO, the scale parameter $t$ can update actively (Fig.~\ref{scale_density} (c)). 
In this condition, however, many $t$'s approach to zero, which means these SAC layers reduce to the plain convolution. 
Thus, we employ LSC to alleviate the degradation of SAC layer. 
Finally, the combination of three components is able to correctly learn the visual scale according to the face data (Fig.~\ref{scale_density} (d)).
Most importantly, we should notice that Table~\ref{ablation} and Fig.~\ref{scale_density} together prove that \textbf{the correct learning of scale parameter from data is the root for the ability of SCAN-CNN}.

\begin{figure}[h]
\centering
\begin{minipage}[t]{0.45\linewidth}
\includegraphics[height=2.8cm]{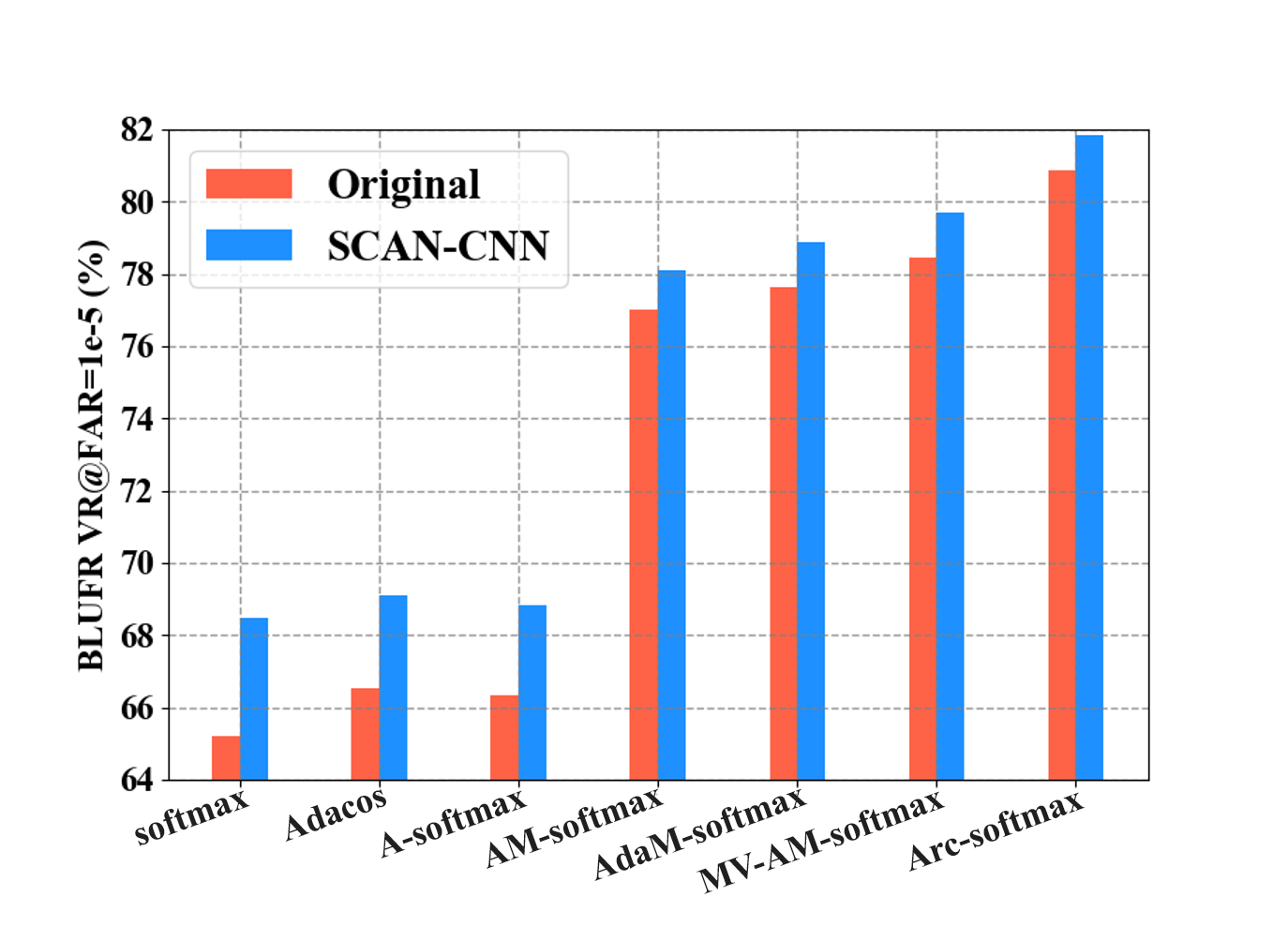}
\subcaption{BLUFR}
\label{BLUFR}
\end{minipage}
\hspace{8pt}
\begin{minipage}[t]{0.45\linewidth}
\includegraphics[height=2.8cm]{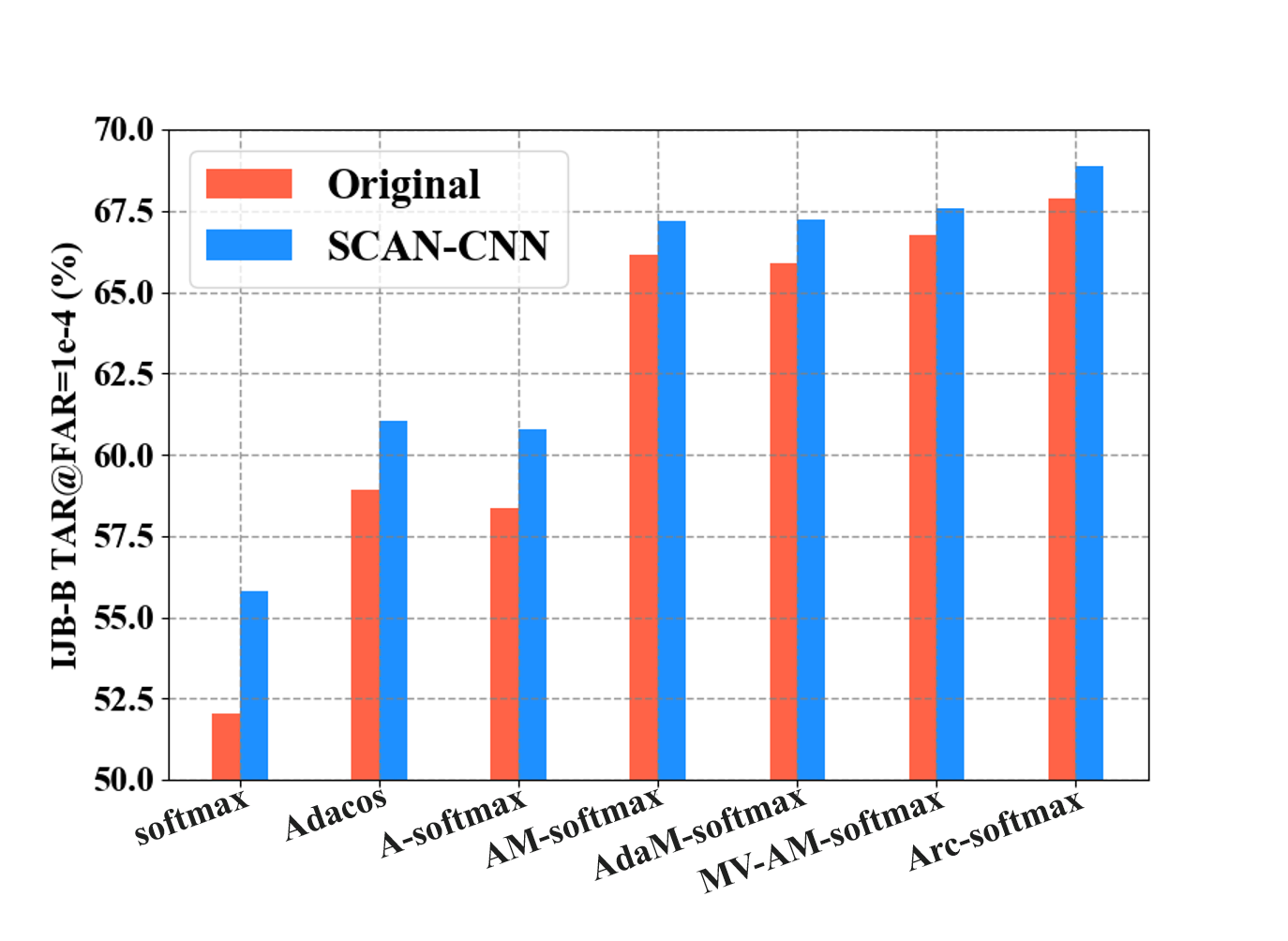}
\subcaption{IJB-B}
\label{IJB-B}
\end{minipage} 
\\
\begin{minipage}[t]{0.45\linewidth}
\includegraphics[height=2.8cm]{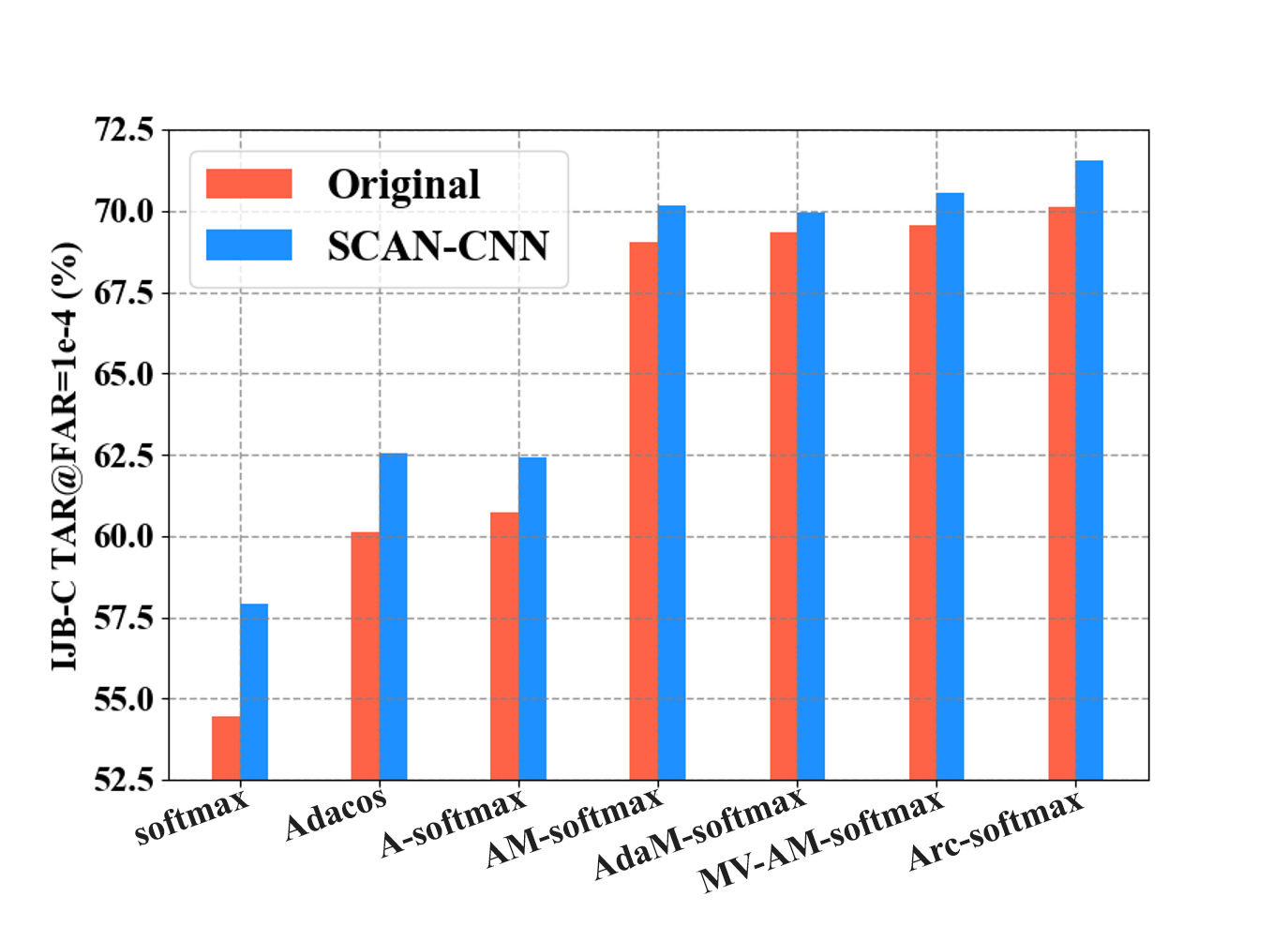}
\subcaption{IJB-C}
\label{IJB-C}
\end{minipage}
\hspace{8pt}
\begin{minipage}[t]{0.45\linewidth}
\includegraphics[height=2.8cm]{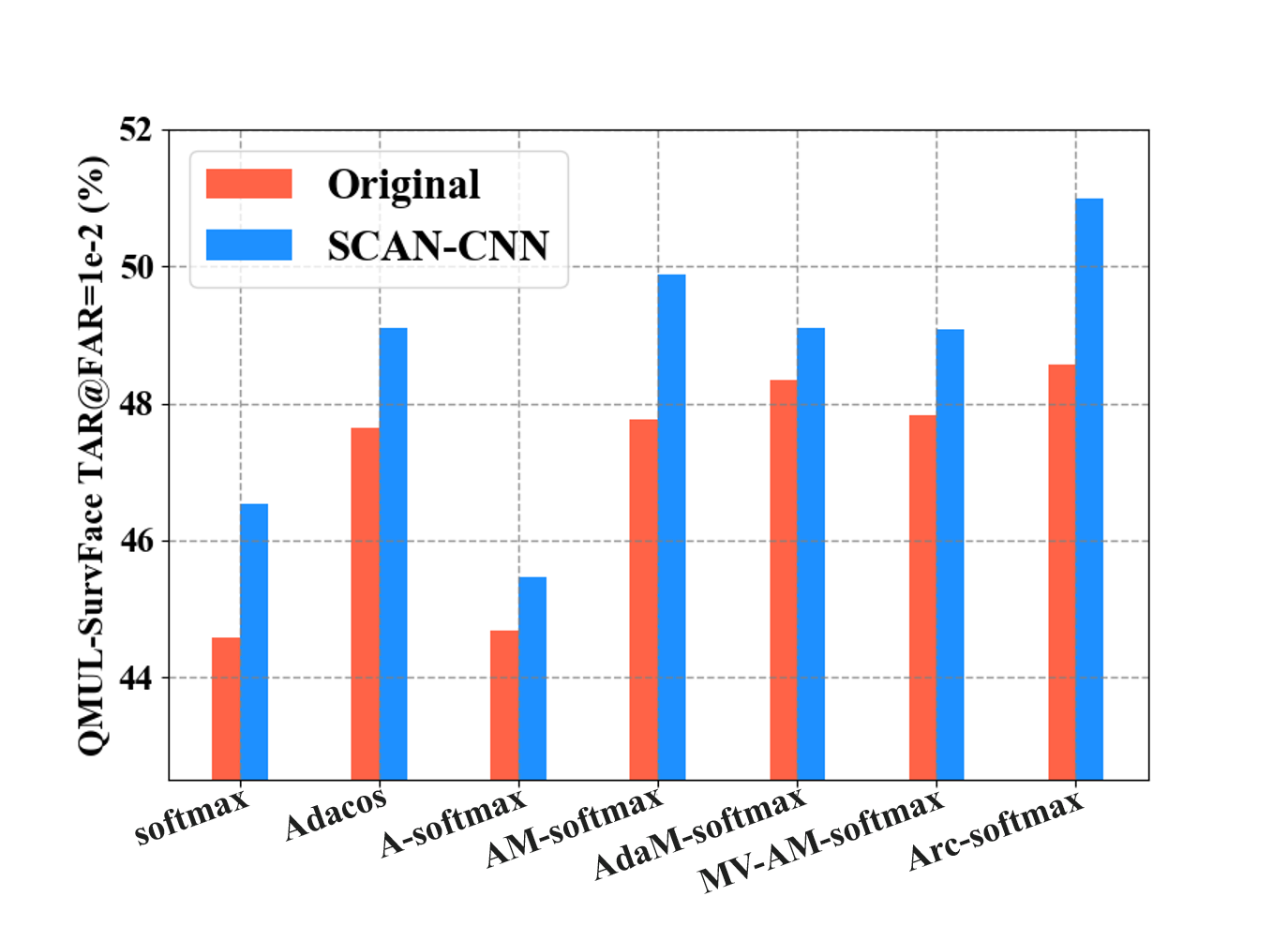}
\subcaption{QMUL-SurvFace}
\label{QMUL-SurvFace}
\end{minipage}
\caption{
SCAN-CNN consistently brings performance gain regardless of loss function.
The red bars denote the plain CNNs, and the blue bars denote SCAN-CNN. }
\label{loss_function}
\vspace{-1em}
\end{figure}

\noindent\textbf{Steady promotion along various loss functions.}
For the training supervision (Eq.~\ref{loss_total}) of SCAN-CNN, the ingredient $\mathcal{L}_{rec}$ can be the off-the-shelf face recognition loss function in literature. 
Here, we try seven representative losses for training SCAN-CNN and the plain counterpart, and test on four test sets (Fig.~\ref{loss_function}). The results show the consistent accuracy improvement compared with the counterparts that without scale attention. Therefore, the advantage of SCAN-CNN certainly comes from the scale attention.

\vspace{-10pt}

\begin{table}[h]
\begin{center}
\centering
\caption{The analysis of attention collaboration.
}
\label{attention}
\resizebox{0.6\linewidth}{!}{
\begin{tabular}{|c|c|c||p{1cm}<{\centering}|p{1cm}<{\centering}|}
\hline
\multirow{2}{*}{Spatial}&
\multirow{2}{*}{Channel}&
\multirow{2}{*}{Scale}&
\multicolumn{2}{c|}{BLUFR}
\\\cline{4-5}&& &$10^{-4}$&$10^{-5}$\\
\hline\hline
-&-&-&83.67&65.19\\
\hline 
-&-&\checkmark&\textbf{85.75}&\textbf{68.48}\\
\hline\hline
\checkmark&-&-&84.77&67.75\\
\hline 
\checkmark&-&\checkmark&\textbf{84.97}&\textbf{68.39}\\ 
\hline \hline 
-&\checkmark&-&84.63&67.00\\
\hline 
-&\checkmark&\checkmark&\textbf{85.73}&\textbf{69.07}\\
\hline \hline

\checkmark&\checkmark&-&84.43&66.21\\
\hline 
\checkmark&\checkmark&\checkmark&\textbf{85.60}&\textbf{68.69}\\

\hline
\end{tabular}}
\end{center}
\vspace{-1em}
\end{table}

\noindent\textbf{Possibility of wide applicability.} 
The behavior in collaboration with existing attentions is critical for applying scale attention to SOTA architectures.
Table~\ref{attention} shows the performance of scale attention, spatial attention, 
channel attention, 
and their combination.
We can see the solo scale attention achieves higher accuracy than the counterparts.
More importantly, both the spatial attention and channel attention can be further enhanced by the combination with scale attention.
This advantage can be taken by many SOTA architectures, since the combination of scale attention with the existing attention mechanisms can be easily implemented in a single CNN.
This is a valuable quality of scale attention for widening its applicability.

\subsection{Advantage over traditional multi-scale}
\label{sec:exp:compare_ms}
In Table~\ref{efficiency}, we make comparison of SCAN-CNN and traditional multi-scale in terms of accuracy and efficiency.
Certain facts can be observed from the results.
Firstly, given the same computational budget, SCAN-CNN (top row) raises the accuracy on BLUFR by 3 percent at false accept rate (FAR) of $10^{-5}$ compared with the single scale baseline (second row).
Secondly, the multi-scale method obtains accuracy raise over baseline by fusing either three (third row) or six (bottom row) CNNs; the more CNNs fused, the more raise gained. However, the multi-scale ensemble also causes the increase of inference cost (rightmost column).
Lastly, SCAN-CNN still outperforms the 6-CNN ensemble in terms of accuracy, and consumes only one-third of the inference cost.
Therefore, SCAN-CNN brings great advantage over multi-scale routine in both accuracy and efficiency.



\begin{table}[h]
\begin{center}
\centering
\caption{
The advantage of SCAN-CNN over the traditional multi-scale routine, in terms of accuracy and efficiency. 
}
\label{efficiency}
\resizebox{0.9\linewidth}{!}{
\begin{tabular}{|c|p{1cm}<{\centering}|p{1cm}<{\centering}|c|}
\hline
\multirow{2}{*}{Method}&
\multicolumn{2}{c|}{BLUFR}&
\multirow{2}{*}{\makecell[c]{Inference\\(ms/image)}}
\\\cline{2-3}&  $10^{-4}$&  $10^{-5}$&\\
\hline\hline
SCAN-CNN&\textbf{85.75}&\textbf{68.48}&\textbf{1.3}\\ 
\hline 
Baseline (single-scale)&83.67&65.19&1.3\\
\hline 
Multi-scale (3-CNN ensemble) &84.16&66.54&2.0\\
\hline 
Multi-scale (6-CNN ensemble) &85.17&68.13&3.8\\

\hline
\end{tabular}}
\end{center}
\vspace{-1em}
\end{table}

\begin{table*}[tp]
\begin{center}
\centering
\caption{
Scale attention boosts the CNNs to obtain accuracy (\%) gain consistently on various benchmarks \textbf{without increase of inference cost}.
The digits before slash are the results on original face images, and after slash are the results on synthetically blurred images.
Under MegaFace, ``Id.'' refers to the face identification rank-1 accuracy with 1M distractors, and ``Veri.'' refers to the face verification TAR at $10^{-6}$ FAR. }
\label{comparison}
\vspace{-5pt}
\resizebox{1\textwidth}{!}{
\begin{tabular}{|c|c|c|c|c|c|c|c|c|c|}
\hline
\multirow{2}{*}{Method}&
\multirow{2}{*}{LFW}&
\multirow{2}{*}{AgeDB-30}&
\multirow{2}{*}{CFP-FP}&
\multirow{2}{*}{CALFW}&
\multirow{2}{*}{CPLFW}&
\multicolumn{2}{c|}{BLUFR}&
\multicolumn{2}{c|}{MegaFace}
\\\cline{7-10}& &&&&&$10^{-4}$&$10^{-5}$&Id.&Veri.\\
\hline\hline 
ResNet-50&99.68 / 84.37 &96.80 / 80.98 &94.61 / 81.51 &94.65 / 81.88 &86.75 / 73.11 &99.20 / 79.84 &94.24 / 59.71 &95.26 / 32.91 &96.36 / 38.89 \\
\hline 
SCAN-ResNet-50 &\textbf{99.72} / \textbf{85.43} &\textbf{97.07} / \textbf{81.82} &\textbf{94.90} / \textbf{82.19} &\textbf{94.80} / \textbf{83.45} &\textbf{87.32} / \textbf{75.13} &\textbf{99.22} / \textbf{82.11} &\textbf{95.01} / \textbf{62.46} &\textbf{95.82} / \textbf{36.39} &\textbf{96.80} / \textbf{41.00} \\
\hline \hline 
SE-ResNet-50&99.70 / 85.23 &97.10 / 81.32 &94.94 / 82.42 &95.08 / 82.09 &87.58 / 74.13 &99.25 / 80.90 &95.33 / 60.53 &96.10 / 34.23 &96.77 / 39.10 \\
\hline 
SCAN--SE-ResNet-50 &\textbf{99.72} / \textbf{86.15} &\textbf{97.18} / \textbf{83.55} &\textbf{95.46} / \textbf{84.63} &\textbf{95.23} / \textbf{83.38} &\textbf{88.10} / \textbf{76.18} &\textbf{99.27} / \textbf{83.78} &\textbf{95.50} / \textbf{63.76} &\textbf{96.35} / \textbf{39.31} &\textbf{96.92} / \textbf{42.73} \\
\hline \hline 
Attention-56&99.71 / 85.62 &97.18 / 81.92 &94.62 / 82.06 &95.25 / 82.30 &87.23 / 75.35 &99.28 / 81.89 &95.29 / 61.12 &96.36 / 35.78 &96.84 / 41.93 \\
\hline 
SCAN-Attention-56 &\textbf{99.73} / \textbf{86.75} &\textbf{97.47} / \textbf{83.95} &\textbf{95.16} / \textbf{84.71} &\textbf{95.28} / \textbf{84.58} &\textbf{87.67} / \textbf{77.17} &\textbf{99.28} / \textbf{83.61} &\textbf{95.54} / \textbf{64.64} &\textbf{96.66} / \textbf{40.55} &\textbf{97.10} / \textbf{45.82} \\
\hline \hline 
ResNet-101&99.73 / 86.05 &97.58 / 82.55 &95.79 / 83.99 &95.30 / 82.85 &88.32 / 75.72 &99.28 / 83.57 &95.21 / 63.22 &96.97 / 39.26 &97.48 / 43.36 \\
\hline 
SCAN-ResNet-101 &\textbf{99.75} / \textbf{86.94} &\textbf{97.72} / \textbf{84.53} &\textbf{96.08} / \textbf{85.23} &\textbf{95.43} / \textbf{84.47} &\textbf{88.56} / \textbf{77.92} &\textbf{99.29} / \textbf{85.79} &\textbf{95.56} / \textbf{65.61} &\textbf{97.27} / \textbf{42.97} &\textbf{97.65} / \textbf{46.51} \\
\hline  \hline 
SE-ResNet-101&99.75 / 86.30 &97.67 / 82.75 &96.11 / 84.09 &95.32 / 83.30 &88.53 / 76.12 &99.30 / 83.30 &95.16 / 64.75 &97.16 / 40.78 &97.69 / 44.55 \\
\hline 
SCAN-SE-ResNet-101 &\textbf{99.75} / \textbf{87.08} &\textbf{97.70} / \textbf{84.63} &\textbf{96.40} / \textbf{85.74} &\textbf{95.45} / \textbf{84.88} &\textbf{88.85} / \textbf{78.15} &\textbf{99.30} / \textbf{86.98} &\textbf{95.52} / \textbf{66.50} &\textbf{97.32} / \textbf{43.19} &\textbf{97.87} / \textbf{47.60} \\
\hline
\end{tabular}}
\end{center}
\vspace{-1em}
\end{table*}

\subsection{Advantage over SOTA CNNs}
\label{sec:exp:compare_plain}

\begin{figure}[tp]
\centering
\includegraphics[width=\linewidth]{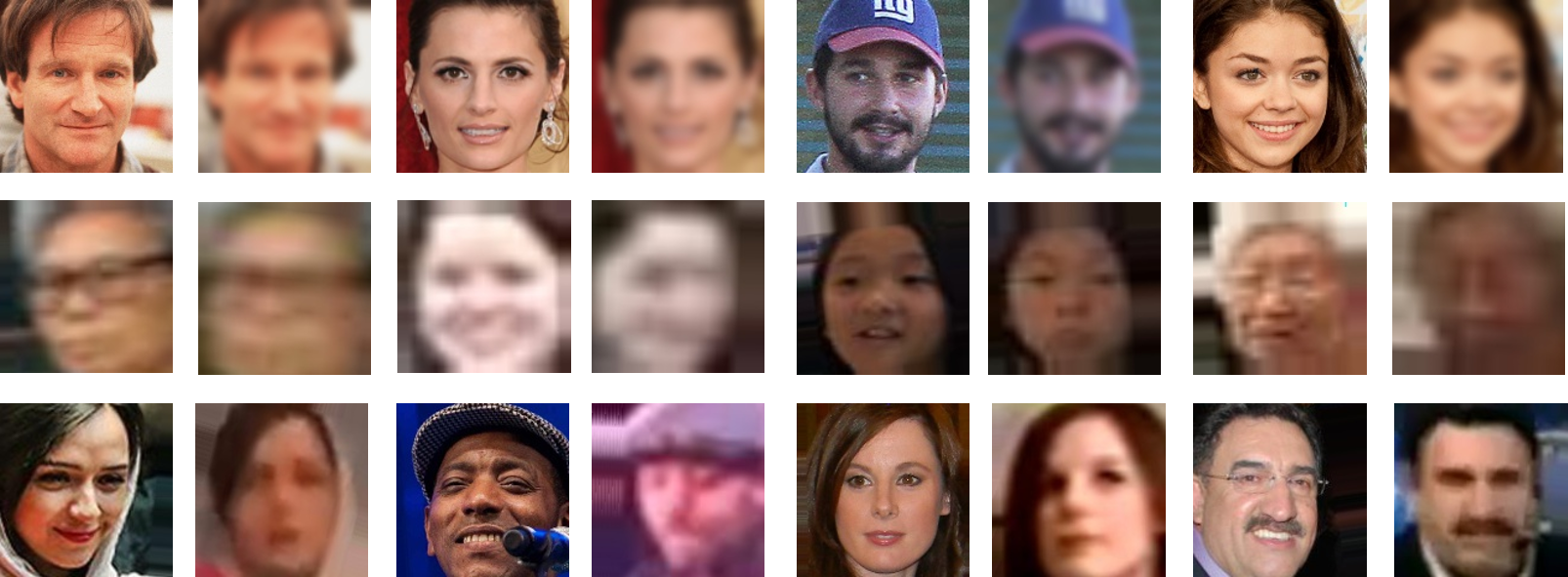}
\caption{Top: examples in MegaFace and synthetic blurring.
Middle: examples in QMUL-SurvFace.
Bottom: clear images and blurry frames in IJB-B and IJB-C.
}
\label{fig:blur_example}
\vspace{-1em}
\end{figure}

\noindent\textbf{General benchmarks and synthetic blurring.} 
The experiment employs five representative CNNs and their SCAN versions as the models, and seven commonly-used test sets as the benchmarks.
In consideration of these test sets mostly containing clear face images, we treat all the test images with synthetic degradation of Gaussian blur (Fig.~\ref{fig:blur_example} Top), in order to verify the advantage of SCAN-CNN against scale variation.
Table~\ref{comparison} shows the result of SCAN-CNN and counterparts on the (original and synthetically blurred) test sets.
From the result, we can find that: 
(1) In each row, the accuracy on blurred face (digits after slash) is much lower than that on original clear face (before slash), 
which indicates the scale variation greatly increases the difficulty of recognition.
(2) In each column, for the blurred face, SCAN-CNN consistently yields evident improvement over the counterparts, which verifies the advantage of SCAN-CNN against the scale variation caused by blurring issue.
(3) For the original clear face, SCAN-CNN still outperforms the counterparts steadily, implying that SCAN-CNN has better robustness to the scale variation caused by different size of facial components (as discussed in Sec.~\ref{sec:disc}).
It is noteworthy that each SCAN-CNN takes the same inference cost compared with its counterpart.

\noindent\textbf{Real-world blurring challenge.} 
QMUL-SurvFace~\cite{cheng2018surveillance} is a real-world surveillance face dataset, which contains many extreme cases of blurry and low-resolution faces (Fig.~\ref{fig:blur_example} Middle). We conduct experiment on this dataset to prove the advantage of SCAN-CNN against real-world blurring challenge. 
Follow the conventional setting, all the SCAN-CNNs and the plain counterparts are pretrained on MS1M-v1c, and then finetuned on the QMUL-SurvFace training set, and finally evaluated on the QMUL-SurvFace test set. 
From Table~\ref{qmul_results}, we can see SCAN-CNN brings obvious improvement in both verification and identification tasks, indicating the robustness and superiority of SCAN-CNN against scale variation in real-world scenario.

\begin{table}[t]
\begin{center}
\label{Finetune}
\caption{Performance improvement on QMUL-SurvFace. 
Note that scale attention promotes the CNNs on extremely-blurry face recognition \textbf{without increase of inference budget}.
}
\label{qmul_results}
\resizebox{0.48\textwidth}{!}{
\begin{tabular}{|c|p{1cm}<{\centering}|p{1cm}<{\centering}|p{1cm}<{\centering}|p{1cm}<{\centering}|p{1cm}<{\centering}|p{1cm}<{\centering}|}
\hline
\multirow{2}{*}{Method}&
\multicolumn{3}{c|}{TPR(\%)@FAR}&
\multicolumn{3}{c|}{TPIR20(\%)@FPIR}
\\ \cline{2-7}&0.1&0.01&0.001&0.3&0.2&0.1\\
\hline\hline 
ResNet-50&72.14&50.19&26.90&25.76&20.81&14.32\\
\hline 
SCAN-ResNet-50 &\textbf{75.96}&\textbf{53.11}&\textbf{28.89}&\textbf{27.50}&\textbf{22.90}&\textbf{15.09}\\
\hline\hline 
SE-ResNet-50&75.19&52.39&28.94&27.40&22.72&14.48\\
\hline  
SCAN-SE-ResNet-50 &\textbf{76.21}&\textbf{54.27}&\textbf{31.99}&\textbf{27.81}&\textbf{23.81}&\textbf{17.10}\\ 
\hline\hline 
Attention-56&74.64&52.73&30.88&26.29&21.66&16.09\\
\hline 
SCAN-Attention-56&\textbf{75.97}&\textbf{53.72}&\textbf{32.74}&\textbf{27.47}&\textbf{22.77}&\textbf{17.23}\\
\hline \hline
ResNet-101&77.47&54.49&34.65&28.64&24.58&18.45\\
\hline 
SCAN-ResNet-101  &\textbf{77.90}&\textbf{56.69}&\textbf{36.08}&\textbf{29.42}&\textbf{25.64}&\textbf{19.07}\\
\hline \hline 

SE-ResNet-101&77.94&55.95&36.15&28.99&25.13&18.56\\
\hline 
SCAN-SE-ResNet-101 &\textbf{78.31}&\textbf{56.26}&\textbf{37.34}&\textbf{29.72}&\textbf{25.52}&\textbf{19.18}\\
\hline
\end{tabular}}
\end{center}
\vspace{-1em}
\end{table}

\begin{figure}[t]
\centering
\begin{minipage}[t]{0.53\linewidth}
\includegraphics[height=3.8cm, width=4.3cm]{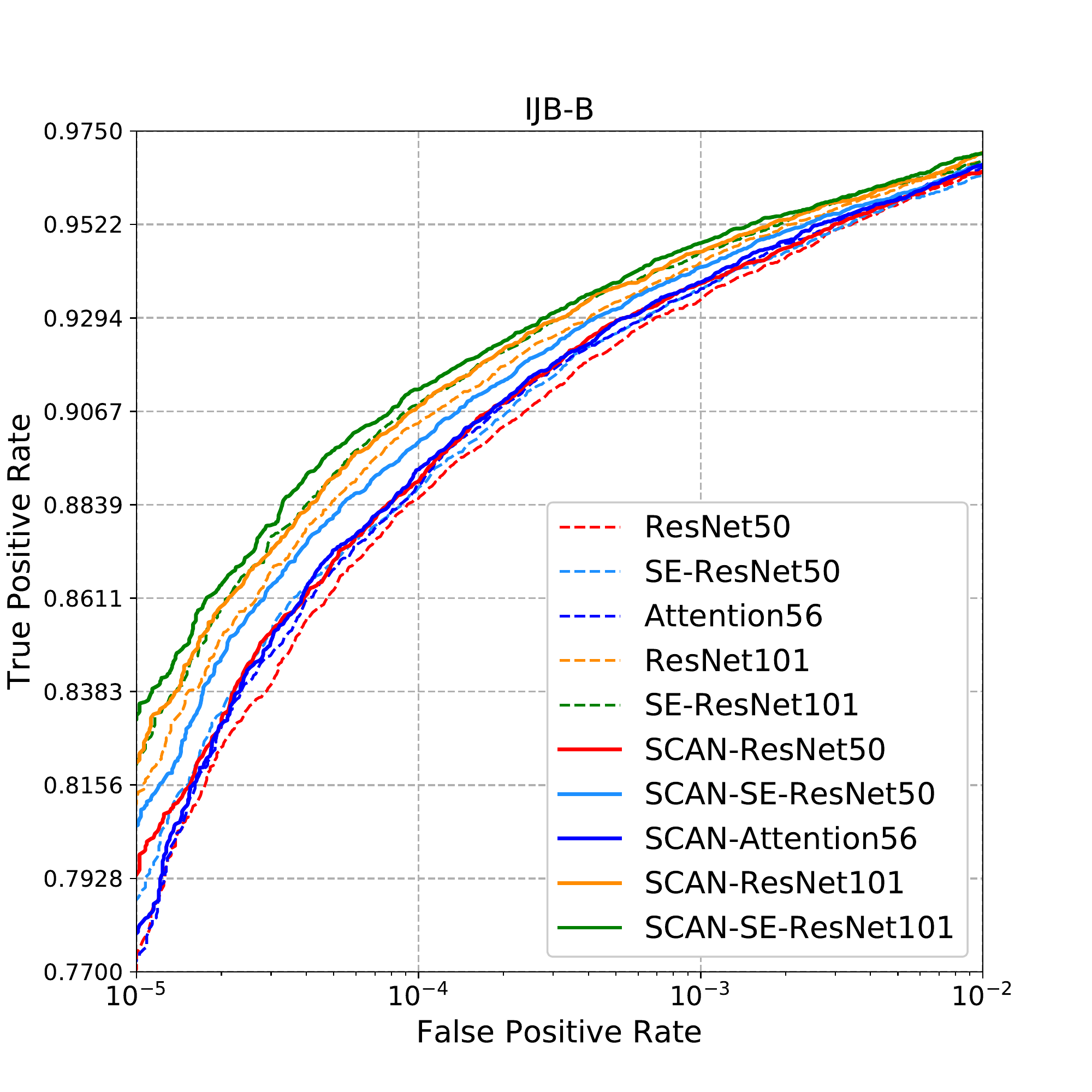}
\subcaption{IJB-B}
\label{IJB-B_curves}
\end{minipage}
\begin{minipage}[t]{0.46\linewidth}
\includegraphics[height=3.8cm, width=4.3cm]{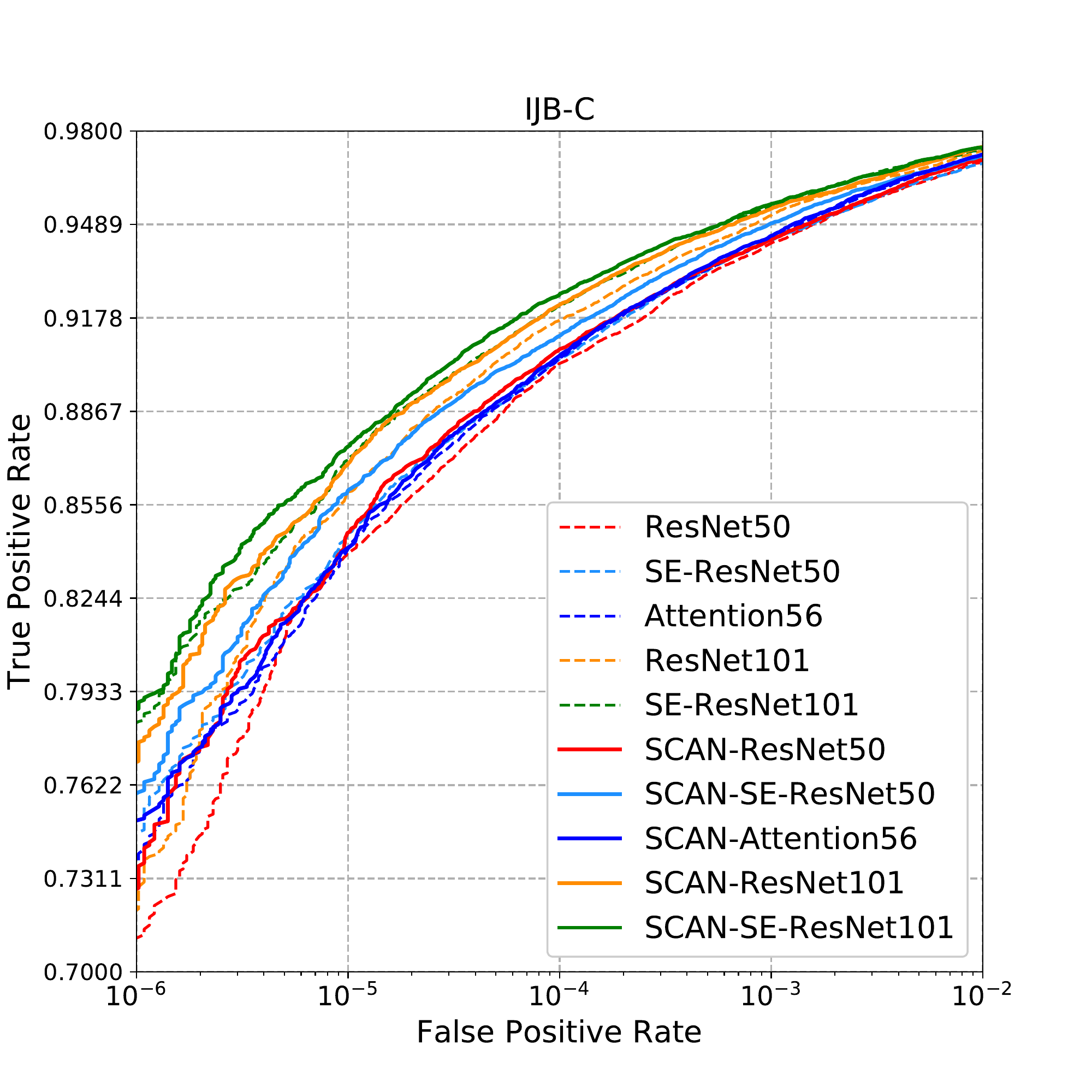}
\subcaption{IJB-C}
\label{IJB-C_curves}
\end{minipage}
\caption{The ROC curves of SCAN-CNNs (solid) and the plain counterparts (dashed). 
Best viewed in color.}
\label{IJB}
\vspace{-10pt}
\end{figure}

\noindent\textbf{Cross image and video.} 
IJB-B~\cite{Whitelam2017IARPAJB} and IJB-C~\cite{Maze2018IARPAJB} involve not only still image but also video frame for face verification.
Therefore, we provide more results on these datasets to compare SCAN-CNN and counterparts dealing with the gap across still image and video frame (Fig.~\ref{fig:blur_example} Bottom). 
As shown in Fig.~\ref{IJB}, 
The SCAN-CNNs consistently achieve higher ROC curves than the plain counterparts, further confirming its robustness to scale variation across image and video.

\section{Conclusion}
\label{sec:conclusion}
In this study, we model the visual scale attention in the CNN regime, 
and propose SCAN-CNN for deep face recognition against scale variation. 
SCAN-CNN is able to learn visual scale from real data, and infer scale-specific representation for robust face recognition.
Compared with multi-scale method, SCAN-CNN improves both accuracy and inference efficiency.
Compared with plain CNNs, SCAN-CNN improves accuracy and robustness against scale variation, and maintain the model parameters and computational cost non-increased in inference.
The experiments shows its advantages and wide applicability.

\appendix
\section{Another Perspective of Understanding the Advantage on Blurry Images}

\begin{figure}[h]
\centering
\hspace{-30pt}
\begin{minipage}[t]{0.4\linewidth}
\centering
\includegraphics[height=2cm]{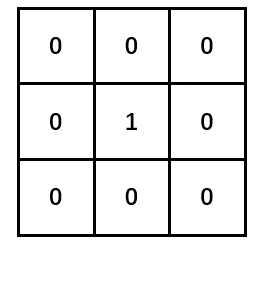}
\subcaption{Dirac}
\label{dirac}
\end{minipage}
\hspace{5pt}
\begin{minipage}[t]{0.4\linewidth}
\centering
\includegraphics[height=2cm]{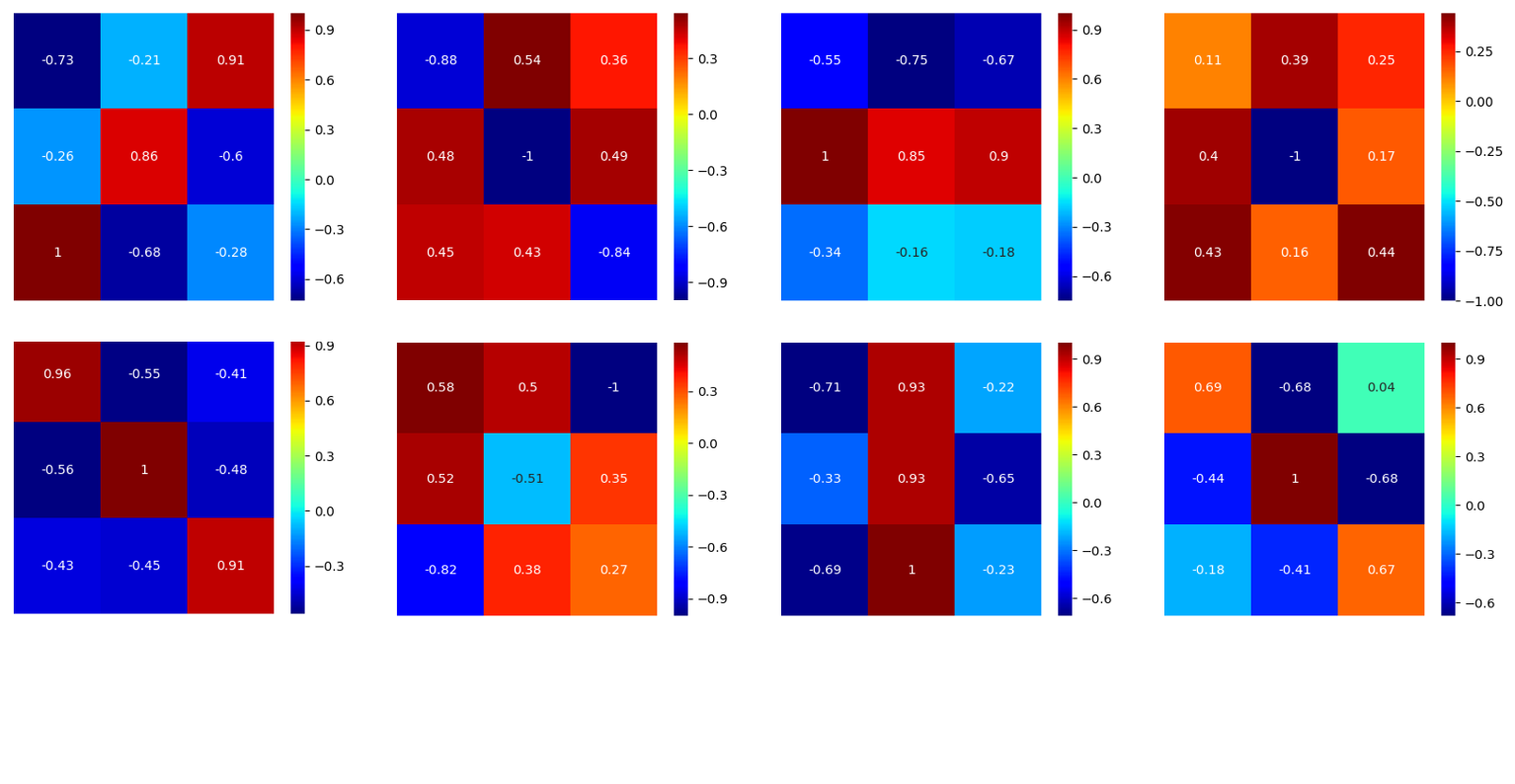}
\subcaption{Learnt filters}
\label{sobel}
\end{minipage}
\caption{
(a) Convolution with the unit-impulse filter (\textit{aka.} the Dirac operator) is identity mapping, which results in the same value of the input itself.
(b) Some examples of normalized visualization of $k_a$, which is estimated by $k_a = iFFT(FFT(k) / FFT(k_i))$.
Zoom in for better view.
}
\label{dirac_sobel}
\vspace{-15pt}

\end{figure}

Here, we provide a new perspective from the position of \textbf{classic image enhancement}, to interpret the effectiveness of SCAN-CNN on blurry face recognition. 
We recall and expand Eq.8 of the main text as follows,
\begin{align}
    f_{out}
    & = 
    k_u \circledast f_{in}
    \label{enhance_1}
    \\
    & =
    (k_s + k_i) \circledast f_{in}
    \label{enhance_2}
    \\
    & =
    (t^{\gamma m}g\circledast k + k_i) \circledast f_{in}
    \label{enhance_3}
    \\
    & = 
    (t^{\gamma m}g\circledast k_a + \mathbb{1}) \circledast k_i \circledast f_{in}
    \label{enhance_4}
    \\
    & = 
    [(t^{\gamma m}g\circledast k_a + \mathbb{1}) \circledast f_{in} ] \circledast k_i
    \label{enhance_5}
    \\
    & =
    (k_{x-pass} \circledast f_{in} ) \circledast k_i,
    \label{enhance_6}
\end{align}
where $\mathbb{1}$ denotes the unit-impulse filter (Fig.~\ref{dirac_sobel} (a)). 
From Eq.~\ref{enhance_4} to Eq.~\ref{enhance_6}, $k$ is decomposed by $k = k_a \circledast k_i$, which leads us to a conclusion that: the input image (or feature map) is processed by some kind of a filter $k_{x-pass}$ before the regular convolution with $k_i$. 
It is important to find out the behavior of $k_a$ and $k_{x-pass}$ with respect to the effect on $f_{in}$.
There are two possibilities.
First, $k_a$ is learnt as a filter with values of opposite signs (possible case similar to the edge enhancement operators such as Sobel, Prewitt, Laplacian \etc), then $k_{x-pass}$ will be a high-pass filter, and the input will be sharpened.
Although $g$ is a Guassian kernel, it should not affect the sharpening, since the opposite signs in $k_a$ guarantee the high-pass differential (such like the Laplacian-of-Gaussian operator).
Second, $k_a$ has same-sign values, then $k_{x-pass}$ will be a low-pass filter, and the input will be smoothed.
The former will be the most actual case in SCAN-CNN (Fig.~\ref{dirac_sobel} (b)), because the training is supervised by the combination of RMO loss and face recognition loss towards robust face representation, and the high-pass filtering is helpful to handling the blurry input.
It is noteworthy that $t$ plays the role of weighting factor here: 
the channel with large $t$ (corresponding to the coarse scale) increases the high-pass proportion in $k_{x-pass}$, enhancing the sharpening effect.

The image-enhancement perspective also verifies the advantage of LSC (local shortcut connection) in SCAN-CNN. LSC provides $k_{x-pass}$ the ready-made unit-impulse component, and thereby facilitates the learning of the high-pass filter. 
If learning without LSC, sole $k_s$ can hardly accomplish the filtering (no matter high- or low-pass) to which the unit impulse is necessary.
The ablation study in the main text shows the profit bought by LSC, and here, this perspective double verifies its advantage.

\section{Heat Equation of Isotropic Diffusion} 
In scale-space theory~\cite{lindeberg1990scale}, the observed signal is regarded as the initial state of ``temperature'' in space, then the scale-space representation can be described by the heat equation of isotropic diffusion. Here, we provide more details about this deduction that mentioned in Section 3.1 of the main text. Specifically, the heat equation of isotropic diffusion can be formulated as:  
\begin{equation}
\partial_{t} F (\cdot;t) =\frac{1}{2} \nabla^{2} F(\cdot;t) =\frac{1}{2} \sum_{i=1}^{D } \partial_{x_{i} x_{i}} F(\cdot;t) ,
\end{equation}
where $F (\cdot;t)$ is the function of temperature (or the image intensity), $D$ is the dimensionality of the spatial coordinates, and $\partial_{x_{i} x_{i}}$ is the second order derivative in the ${i}$-th dimension. 

The equation describes that temperature is isotropically diffusing in space with the increasing time $t$. 
In scale-space theory, the visual scale corresponds to the ``time'' variable, and the scale-space representation gradually loses the high frequency information with the increase of visual scale. 
The linear solution of scale-space representation can be obtained in a closed form by the Gaussion convolution, 
\begin{equation}
F(\cdot; t) = g(\cdot; t) \circledast f(\cdot), 
\label{scale_space_representation}
\end{equation} 
where $f$ is the observed signal, and $g$ is the Gaussian kernel with the variance associated with the scale $t$.

\section{Non-monotonicity of Normalized Derivative}
The normalized derivative provides two valuable properties for the automatic scale selection in the scale-space representation, which has been discussed in Section 3.2 of the main text. Here, the formal proof of the non-monotonicity of normalized derivative is given as follow. 
Likewise, we take an example of a sinusoidal signal $f(x) = sin(\omega x)$. 
According to Eq.~\ref{scale_space_representation}, the scale-space representation of $f(x)$ can be computed by $F(x;t) = e^{-\omega^2 t/2} sin(\omega x)$, and its amplitude of the $m$-th order normalized derivative can be written as: 
\begin{equation}
    F_{\xi^m, max}(t) = t^{m\gamma} \omega^m e^{-\omega^2 t/2}, 
\end{equation}
where the subscript $\xi^m$ denotes the $m$-th order normalized derivative $\partial_{\xi^m}$, with $\partial_{\xi} = t^\gamma \partial_x$, and the subscript $max$ denotes the amplitude of $F_{\xi^m}$. 
We take the derivative of $F_{\xi^m, max}(t)$ over scale $t$ as an example, which can be computed by
\begin{align}
    \frac{\partial F_{\xi^m, max}(t) }{\partial t} 
    & = {m\gamma t^{m\gamma - 1} \omega^m e^{-\omega^2 t/2} - 
    t^{m\gamma}\frac{\omega^2}{2} \omega^m e^{-\omega^2 t/2}} 
    \nonumber \\
    & = {\left (m\gamma t^{m\gamma - 1} - t^{m\gamma}\frac{\omega^2}{2} \right)} \omega^m e^{-\omega^2 t/2} 
    \nonumber \\
    & = {\left (\frac{ m\gamma} {t} - \frac{\omega^2}{2} \right)}t^{m\gamma}  \omega^m e^{-\omega^2 t/2}
    \nonumber \\
    & = {\left (\frac{2 m\gamma - \omega^2t } {2t} \right)}t^{m\gamma}  \omega^m e^{-\omega^2 t/2}. 
\end{align}

We can observe that the derivative over scale is initially larger than zero and gradually falls below zero with the increasing scale, confirming the non-monotonicity of normalized derivative. 
The unique maximum is attained at the scale $t = 2m\gamma / \omega^2$, which precisely reflects the real scale of the signal.


\section{Extension of Normalized Derivative}
The automatic scale selection~\cite{lindeberg1998feature} extends the normalized derivatives to the normalized derivative polynomial for various feature detection tasks. In our method, we consider that the convolution kernel of CNN can be decomposed by the linear combination of the finite discrete operators of derivatives with different orders, which can be formulated by

\begin{equation}
d(F)=\sum_{i=1}^{I} c_{i} \prod_{j=1}^{J} F_{{x}^{\alpha_{i j}}}, 
\end{equation}
where 
$ F_{{x}^{\alpha_{i j}}}$ is the $\alpha_{i j}$-th order derivatives over space, 
$i$ and $j$ are the indexes of polynomial terms and subterms, 
$c_{i}$ is the coefficient of the $i$-th term, and $d(F)$ denotes the entire derivative polynomial. Then, the normalized derivative polynomial is defined as:  
\begin{equation}
d_{norm}(F)=  t^{ M\gamma} d(F), \quad
with \quad M = \sum_{j=1}^{J}\left|\alpha_{i j}\right|,
\end{equation}
which is another shape of the kernel $t^{\gamma m} k(\cdot)$ regarded from the perspective of normalized derivative polynomial. Note that, no matter from either perspective, $t^{\gamma m} k(\cdot)$ is  updated in the data-driven manner according to the training objective.



\section{Efficient Training with RMO}
\begin{figure}[h]
    \centering
    \includegraphics[scale=0.4]{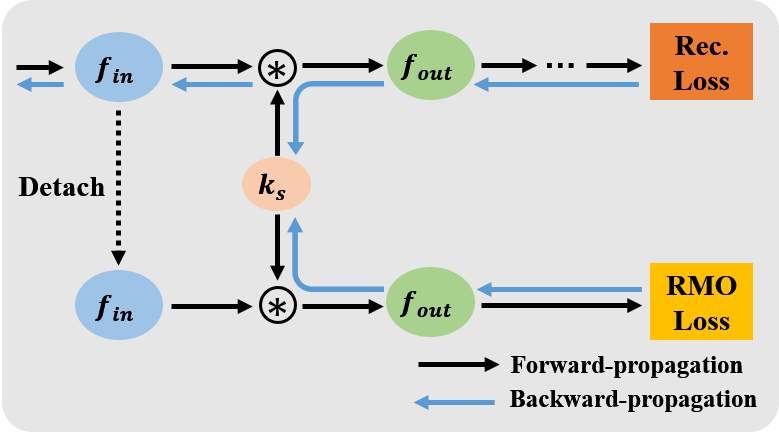}
    \caption{The design details of efficient training with RMO in a scale attention conv layer. The gradients with respect to $k_s$ are derived from the combination of RMO and recognition loss. }
    \label{RMO}
\end{figure}
As mentioned in Section 4.2 of the main text, the RMO loss takes effect only for the corresponding layer. In this situation, the original implementation of backward-propagation in PyTorch is suboptimal in terms of efficiency, since the redundant gradients will be calculated in the backward-propagation of the previous layers. 
As shown in Fig.~\ref{RMO}, we address this issue by detaching the corresponding nodes (the blue ellipse) from the whole computing graph and cutting off the redundant gradient flow. 
By doing so, the time cost reduces from 24 to 10 hours for training a network such as SCAN-ResNet-18.

\section{Convergence of SCAN-CNN}

\begin{figure}[h]
    \hspace{-10pt}
    \includegraphics[scale=0.4]{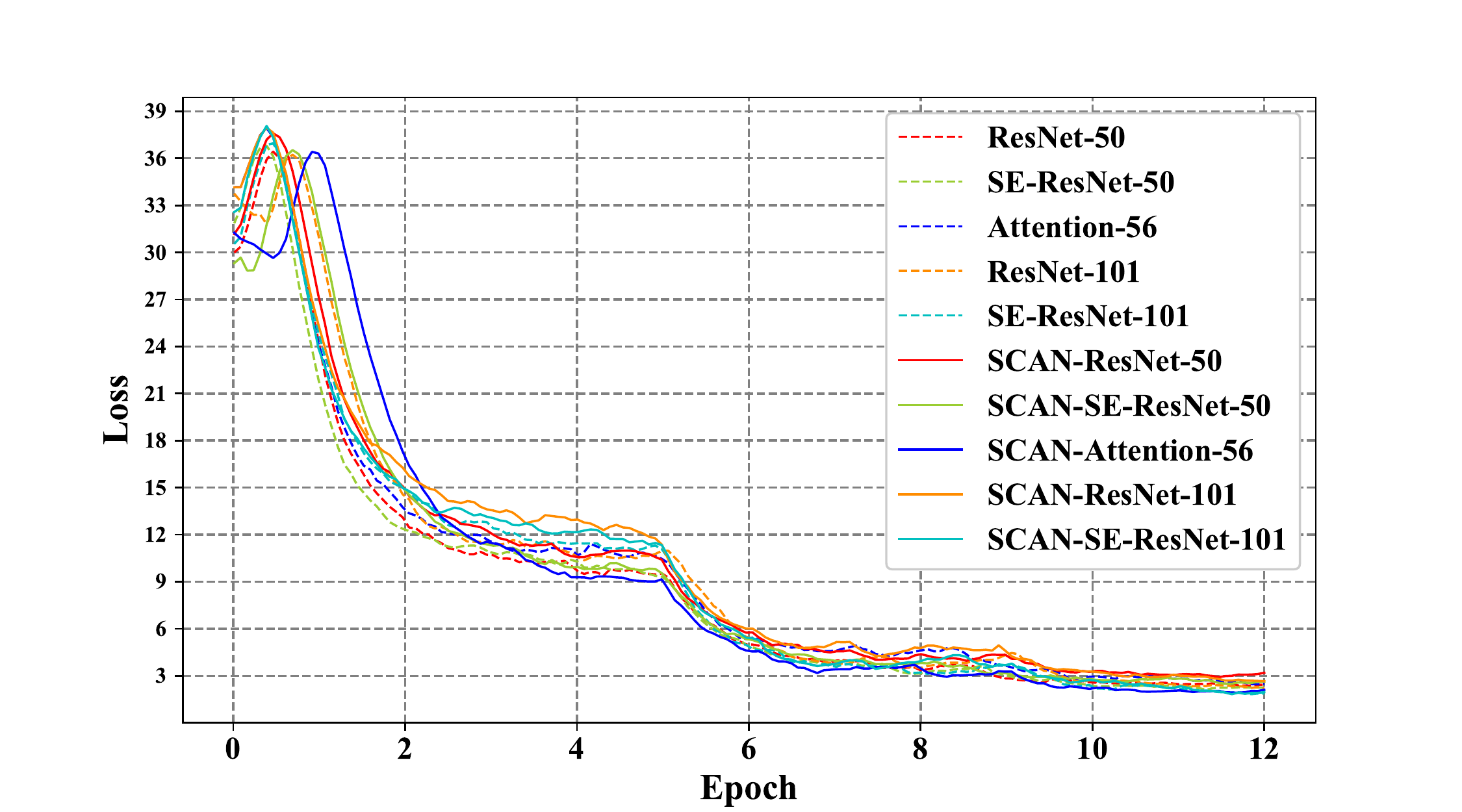}
    \caption{The training loss curves of SCAN-CNNs and their counterparts. 
Best viewed in color.}
    \label{loss_curve}
\end{figure}

We provide the training loss curve of SCAN-CNN to confirm its stability in terms of training convergence.
As shown in Fig.~\ref{loss_curve}, the SCAN-CNNs converge steadily as their plain counterparts do. 
This is also an evidence of the correct learning of scale parameter $t$, because the loss value of SCAN-CNN includes the RMO loss term.
If $t$ fails to learn the correct scale, the RMO loss will not decrease.

{\small
\bibliographystyle{ieee_fullname}
\bibliography{egpaper_final}
}

\end{document}